%% file: TEAI_paper.tex
\documentclass[]{TEAI}

\usepackage{helvet}

\usepackage{amsmath} 
\usepackage{mathtools}
\usepackage{natbib}
\usepackage{graphicx}
\usepackage{subcaption} 
\input{math_commands.tex}

\usepackage[toc,page,header]{appendix}
\usepackage[utf8]{inputenc} 
\usepackage[T1]{fontenc}    
\usepackage{hyperref}       
\usepackage{url}            
\usepackage{booktabs}       
\usepackage{lmodern}        
\usepackage{amsfonts}       
\usepackage{nicefrac}       
\usepackage{microtype}      
\usepackage{wrapfig}

\usepackage{amssymb}  
\usepackage{fontawesome}  
\usepackage{url}  

\usepackage{titletoc}

\usepackage{tikz}  
\usepackage{comment}  
\usepackage{tabularx}  
\usepackage{booktabs}  

\usepackage{minitoc}

\usepackage{booktabs}
\usepackage{array}
\usepackage{etoolbox}

\definecolor{lightblue}{RGB}{200, 230, 255}  
\definecolor{headerblue}{RGB}{150, 200, 255} 

\usepackage{pgfplots}
\usepackage[utf8]{inputenc} 
\usepackage[T1]{fontenc}    
\usepackage{hyperref}       
\usepackage{url}            
\usepackage{booktabs}       
\usepackage{amsfonts}       
\usepackage{nicefrac}       
\usepackage{microtype}      
\usepackage{graphicx}
\usepackage{float}
\usepackage{comment}
\usepackage{multirow} 
\usepackage{amsmath} 
\usepackage{makecell} 
\usepackage{siunitx}  
\usepackage{tikz}
\usepackage{pgf-pie} 
\usepackage{subcaption}
\usepackage{wrapfig}
\usepackage[export]{adjustbox}

\usepackage{ragged2e}      
\usepackage{tabularx}       
\usepackage{array}          
\usepackage{caption}        
\usepackage{enumitem}
\usepackage{pifont}
\usepackage[hang,flushmargin]{footmisc} 
\usepackage{CJKutf8}

\usepackage[ruled,vlined]{algorithm2e}

\usepackage{tcolorbox}
\usepackage{environ}

\usepackage{tcolorbox}    
\tcbuselibrary{breakable}  
\tcbuselibrary{skins}      

\usepackage{tabularx}
\usepackage{listings}

\definecolor{mygreen}{RGB}{76, 200, 80}  
\definecolor{mygray}{RGB}{240, 240, 240} 
\definecolor{myred}{RGB}{244, 67, 54}     

\theoremstyle{plain}

\newtheorem{proposition}{Proposition}

\theoremstyle{definition}

\theoremstyle{remark}

\newcommand{\Heff}{H_{\mathrm{eff}}}
\newcommand{\Hcov}{H_{\mathrm{cov}}}
\newcommand{\Dkl}{D_{\mathrm{KL}}}

\tcbset{
  prompt/.style={
    width=\linewidth,
    top=8pt,
    bottom=4pt,
    colback=SeaGreen!10!CornflowerBlue!10,
    colframe=black,
    colbacktitle=black,
    enhanced,
    center,
    attach boxed title to top left={yshift=-0.1in,xshift=0.15in},
    boxed title style={boxrule=0pt,colframe=white,},
  }
}

\newtcolorbox{PromptBox}[2][]{prompt,breakable,title=#2,#1}

\newtcolorbox{turnbox}[1][]{
  colback=blue!5!white, colframe=blue!75!black, boxrule=0.5pt, arc=3pt, left=2pt, right=2pt, top=2pt, bottom=2pt, enhanced, width=0.25\textwidth,
  #1
}

\title{\textsc{TurnOPD}: Making On-Policy Distillation Turn-Aware\\ for Efficient Long-Horizon Agent Training}

\author{
    Yuhang Zhou\textsuperscript{1,2},
    Kai Zheng\textsuperscript{2,*,$\dagger$},
    Haoling Li\textsuperscript{2}, 
    Dengyun Peng\textsuperscript{1},
    Can Xu\textsuperscript{2,$\dagger$}, 
    Jingjing Chen\textsuperscript{1,$\dagger$}
}

\affiliation[1]{\mbox{Fudan University}} 
\affiliation[2]{\mbox{Tencent Hunyuan}}

\correspondence{\textcolor{seedblue}{ralph.yh.zhou@gmail.com, \{kevinezheng,leocaxu\}@tencent.com, chenjingjing@fudan.edu.cn}}

\long\def\teaisetabstract#1{\gdef\abstractlist{{\abstractinfont #1}}}
\RenewEnviron{abstract}{\expandafter\teaisetabstract\expandafter{\BODY}}

\begin{document}
\input{sections/abstract}

\maketitle
\renewcommand{\thefootnote}{}
\footnotetext{$^*$Project lead.\\$^\dagger$Corresponding authors.}
\renewcommand{\thefootnote}{\arabic{footnote}}


\input{sections/introduction}

\input{sections/related_work}

\input{sections/preliminaries}

\input{sections/diagnosis}

\input{sections/method}

\input{sections/experiments}

\input{sections/ablation}

\input{sections/conclusion}

\clearpage

\bibliographystyle{plainnat}
\bibliography{tasc}

\input{sections/appendix}






\end{document}

%% file: math_commands.tex
\usepackage{amsmath,amsfonts,bm}

\def\eqref#1{equation~\ref{#1}}

\def\1{\bm{1}}

\DeclareMathAlphabet{\mathsfit}{\encodingdefault}{\sfdefault}{m}{sl}
\SetMathAlphabet{\mathsfit}{bold}{\encodingdefault}{\sfdefault}{bx}{n}

\newcommand{\E}{\mathbb{E}}

\newcommand{\KL}{D_{\mathrm{KL}}}



%% file: sections/abstract.tex
\begin{abstract}
On-policy distillation (OPD) trains a student policy by matching a stronger teacher on the student's own trajectories, offering a promising framework for language agent training. However, its application to long-horizon agentic tasks remains insufficiently explored. We identify two key inefficiencies in vanilla agent OPD: (1) full-horizon rollouts often waste wall-clock resources on tail turns that provide weak and noisy KL supervision, and (2) trajectory-level KL objectives concentrate most of the loss on shallow tokens, leaving deeper decision turns under-trained once initial behaviors are aligned. To address these challenges, we propose \textbf{TurnOPD}, a turn-level budgeting strategy for efficient on-policy distillation of long-horizon agents. TurnOPD consists of two budget controllers: adaptive rollout-depth budgeting, which uses probe-based turn statistics to determine rollout length, and progressive turn-normalized loss budgeting, which gradually shifts KL weighting from token-level to turn-balanced supervision. Experiments on ALFWorld, WebShop, and Multi-Hop Search with task-specialized teacher models show that TurnOPD achieves superior validation accuracy under equal wall-clock training budgets and advances the accuracy--time frontier beyond vanilla OPD.
\end{abstract}

%% file: sections/introduction.tex
\section{Introduction}
\label{sec:intro}

Language models are increasingly deployed as agents for planning, tool use, and environment interaction~\citep{gao2025survey, hu2026owl, luo2025large, luo2025agentmath, zhou2026offseeker}. On-policy distillation (OPD) offers a promising training framework for such agentic models~\citep{lu2025onpolicydistillation, li2026rethinking}, where a student samples rollouts and a stronger teacher supervises via a reverse-KL objective at visited states. This keeps supervision on-policy and provides dense feedback while avoiding sparse reward signals.

Yet, applying OPD to long-horizon agent tasks is challenging because these are more than simple sequence-generation problems~\citep{zhong2026sod,wang2026tcod}. Agent rollouts involve multiple turns, tool calls, environmental shifts, and state changes. Early decisions impact all subsequent states, and later turns often contain key but infrequent decisions, so token-level feedback alone may not provide proper supervision for all crucial points.

\begin{figure}[t]
\centering
\includegraphics[width=0.98\linewidth]{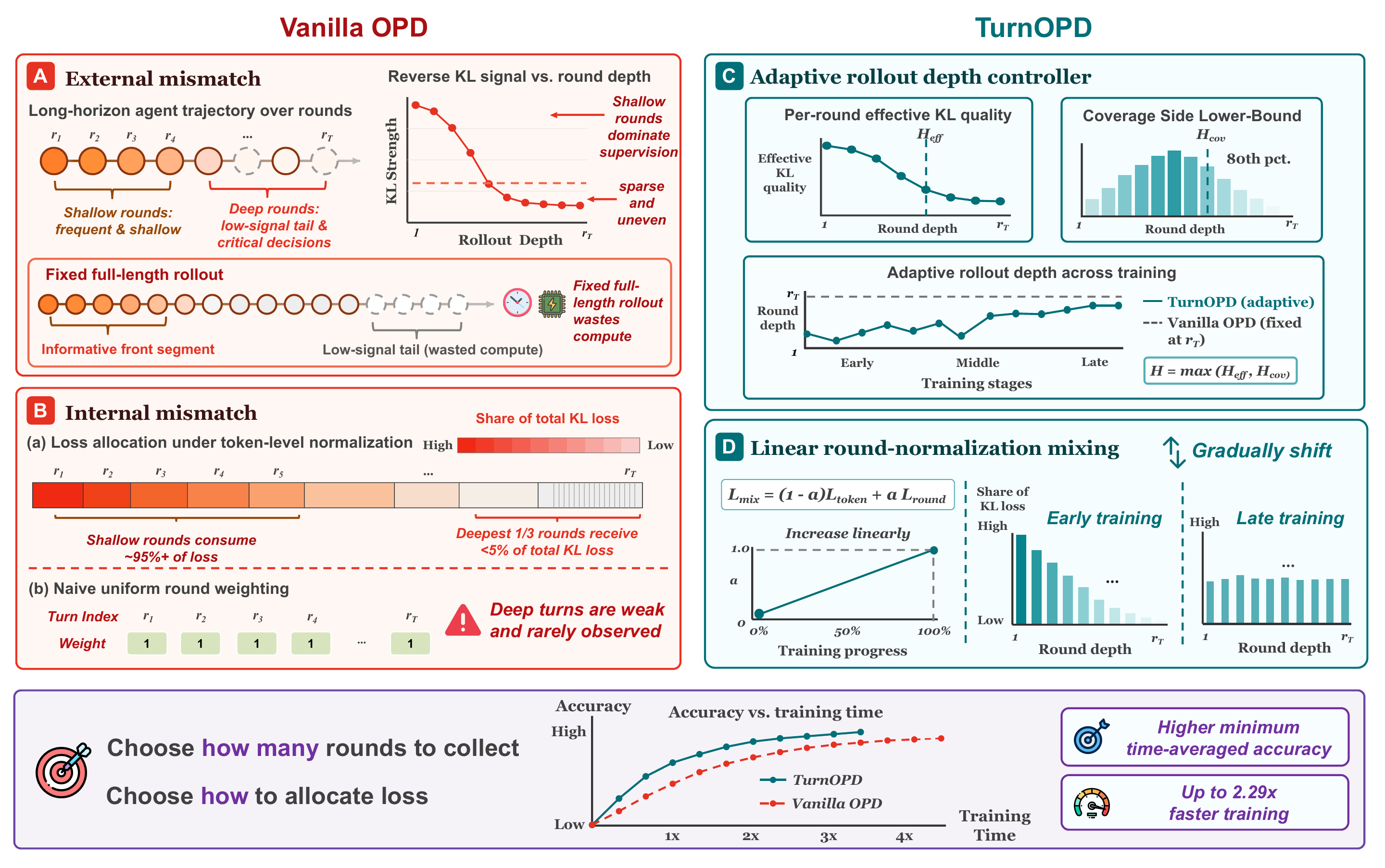}
\caption{A turn-aware perspective on agent OPD. Standard OPD fixes the rollout depth and averages KL across the full trajectory, often wasting resources on low-value tail turns and over-focusing loss on shallow tokens. TurnOPD budgets both rollout depth and KL aggregation based on turn-level statistics, providing a more balanced supervision signal.}
\label{fig:intro-tasc-overview}
\vspace{-0.3cm}
\end{figure}

We systematically analyze OPD in long-horizon agent tasks, asking: \textit{How are supervision signals and optimization budgets allocated along the interaction trajectory, and with what consequences?} Our diagnosis identifies two main problems: \textbf{(1) Per-turn KL distribution:} The reverse-KL signal is heavily skewed toward early turns, changes over training, and becomes less informative for deeper steps. Student and teacher outputs often converge as more context is self-generated, reducing observed KL even when meaningful disagreements persist. \textbf{(2) Loss budget:} The optimizer's loss is not equitably distributed—shallow turns are overrepresented and carry higher KL, so they dominate the loss, while deep turns contribute little. For example, only $3.6$--$4.5\%$ of the KL loss is assigned to the deepest third of turns on ALFWorld~\citep{shridhar2021alfworld}, and $11$--$13\%$ on Multi-Hop Search. While strict turn-level normalization can mitigate this, it risks over-weighting deep, low-data turns. Overall, standard KL supervision and normalization along full trajectories are insufficient.

These issues stem from two mismatches: (1) \textbf{External mismatch:} Fixed rollout depth ignores that both correction signal and survivor count vary markedly with turn. Expending compute up to the maximum horizon wastes effort on steps with little reliable signal. (2) \textbf{Internal mismatch:} Trajectory-level normalization gives uniform token weights, concentrating the KL signal on easy, shallow turns and starving deep, informative ones once the model learns basic interaction patterns.

To address these, we propose \textbf{TurnOPD}, a turn-level budgeting strategy for efficient on-policy distillation of long-horizon agents. TurnOPD consists of two budget controllers. The \textbf{adaptive rollout-depth} controller dynamically selects rollout length via periodic probes, guided by survivor-weighted KL and coverage thresholds. The \textbf{progressive turn-normalized loss} controller gradually transitions KL loss aggregation from trajectory-level to turn-balanced, increasing the weight of deep turns.

We evaluate TurnOPD on three representative multi-turn agent benchmarks: embodied planning (ALFWorld), web navigation (WebShop)~\citep{yao2022webshop}, and Multi-Hop Search (including PopQA~\citep{mallen2023popqa}, NQ~\citep{kwiatkowski2019natural}, 2WikiMultiHopQA~\citep{ho2020twowiki}, HotpotQA~\citep{yang2018hotpotqa}). TurnOPD improves minimal-time Avg@4 across all tasks and models. For example, on ALFWorld-1.7B, it increases Same-Step Avg@4 from $83.0$ to $86.3$ and cuts 100-step wall time from $4.42$h to $1.93$h. On WebShop, both accuracy and wall time (from $1.57$h to $1.24$h) are improved.
Our ablations further support the decomposition. Adaptive depth is the main efficiency lever: on ALFWorld-1.7B, it nearly halves wall time but does not by itself solve the loss-allocation problem, while linear KL blending improves Same-Step Avg@4 by reallocating supervision toward deeper turns. Combining both gives the best accuracy--time tradeoff, and coverage-floor studies show that the controller can be tuned smoothly between conservative and aggressive rollout horizons.

In summary, our contributions are as follows:
\begin{itemize}[leftmargin=1.5em,itemsep=2pt,topsep=3pt]
\item We provide a turn-level analysis of OPD for long-horizon agents, revealing the general distribution of supervision and optimization signals across decision turns.
\item We formalize one contamination-compression mechanism for the mismatch phenomenon of OPD in long-horizon agent tasks, and demonstrate the existence of a potentially optimal rollout horizon.
\item We propose \textbf{TurnOPD}, which combines adaptive rollout-depth budgeting with progressive turn-normalized loss budgeting.
\item We demonstrate that TurnOPD achieves the best Least-Time accuracy on tested benchmarks, advancing the accuracy--time frontier. Notably, TurnOPD achieves comparable or even better performance with up to 2.29$\times$ faster training.
\end{itemize}

%% file: sections/related_work.tex
\section{Related Work}
\label{sec:related}

\textbf{On-policy distillation.}
Early work such as MiniLLM and GKD introduced on-policy distillation (OPD) recipes for LLMs: MiniLLM applies reverse-KL training on student-generated samples \citep{gu2024minillm}, while GKD mixes on- and off-policy data and explores various divergence objectives \citep{agarwal2024opd}. Later work viewed OPD as KL-constrained policy optimization, with the teacher--student log-ratio as a token-level reward \citep{lu2025onpolicydistillation,yang2026gopd}, inspiring methods that stabilize target distributions, consider teacher entropy, or relax imitation constraints \citep{jang2026veto,jin2026entropy,ko2026reopold,zhao2026self}. Diagnostic studies point out that dense supervision on long trajectories can be unreliable \citep{li2026rethinking,zhang2026full}. Still, most work regards each example as a single response, whereas long-horizon agents require finer granularity. TCOD introduces a curriculum over trajectory length \citep{wang2026tcod}, and SOD reweights distillation at the step level for tool-assisted reasoning \citep{zhong2026sod}, both highlighting that OPD for agents demands more than a flat sequence objective. Our work complements these by analyzing rollout and gradient allocation across turns, and adapting rollout depth and loss accordingly.

\textbf{Long-horizon agent tasks.}
Agent benchmarks have shifted from controlled interaction
environments to realistic computer-use tasks. ALFWorld links text planning with
embodied environments \citep{shridhar2021alfworld}, while WebShop studies grounded
e-commerce navigation \citep{yao2022webshop}. Later web and GUI
benchmarks broaden the setting to real websites and desktop or mobile
interfaces, including Mind2Web, WebArena, OSWorld, and AndroidWorld
\citep{deng2023mind2web,zhou2023webarena,xie2024osworld,rawles2025androidworld}.
Recent evaluations further target practical work: SWE-bench measures repository
editing from GitHub issues \citep{jimenez2023swebench}, Terminal-Bench evaluates
command-line workflows \citep{merrill2026terminalbench}, BrowseComp measures
persistent web research \citep{wei2025browsecomp}, and TheAgentCompany,
GDPval, and ClawBench move toward professional or everyday online tasks
\citep{xu2024agentcompany,patwardhan2025gdpval,zhang2026clawbench}. Across
these settings, the common structure is not a flat response but a stateful
interaction trace with external observations, changing action validity, and
changing sets of active trajectories across turns.

%% file: sections/preliminaries.tex
\section{Preliminaries: On-Policy Distillation}
\label{sec:prelim}
\textbf{Multi-turn agent interaction.}
We consider a student agent that interacts with an external environment over
multiple turns. At turn $t$, the agent observes $o_t$, conditions on the full
history
\[
h_t=(x,o_1,r_1,o_2,r_2,\ldots,o_t),
\]
and generates a model response $r_t\sim\pi_\theta(\cdot\mid h_t)$. The response
may contain reasoning text and tool arguments. The
environment then maps the executable part of $r_t$ to the next observation
$o_{t+1}$. A rollout is therefore an interaction trace
\[
\tau=(x,o_1,r_1,o_2,r_2,\ldots,o_T,r_T),
\]
which terminates when the task is completed, the environment returns a terminal
state, or a maximum horizon is reached. This turn structure matters because an
early student action changes later observations and thus changes the future
contexts on which the teacher will be queried.

\textbf{Multi-turn OPD.}
We use on-policy distillation (OPD) \citep{lu2025onpolicydistillation}: the student
samples its own rollout $\tau\sim\pi_\theta(\cdot\mid x)$ for a prompt
$x\sim p_{\mathrm{data}}$, and a frozen teacher $\pi_T$ provides token-level
supervision on the student-visited prefixes. For token position $i$ in the
concatenated model responses, let $s_i$ denote the full prefix before that token,
including previous observations, previous student responses, and the current
partial response. This differs from offline distillation because the teacher is
queried on histories induced by the current student rather than only on
teacher-generated demonstrations.

For a response-token mask $m_i$, we use reverse KL as the OPD objective:
\begin{equation}
\mathcal{L}_{\mathrm{OPD}}(\theta)
=
\mathbb{E}_{x,\tau}
\left[
\frac{1}{\sum_i m_i}
\sum_{i=1}^{L}
m_i\,
\Dkl\!\left(
\pi_\theta(\cdot\mid s_i)\,\|\,\pi_T(\cdot\mid s_i)
\right)
\right].
\label{eq:opd-rkl}
\end{equation}

%% file: sections/diagnosis.tex
\section{Diagnosis: Signal Structure in Agent OPD}
\label{sec:diag}

Long-horizon agent OPD still optimizes a teacher-matching loss over a generated
sequence, but the sequence is also an interaction trace.
This section gives a turn-resolved diagnosis of \textbf{how OPD supervision behaves in
long-horizon agent tasks}, focusing on where the teacher signal appears, how
reliably it separates outcomes, and how much loss budget it receives.


\subsection{Raw Turn-Level KL Exposes Non-Uniform Supervision}
\label{sec:diag:surv}

Prior OPD and agent-distillation work has emphasized
long-trajectory exposure and the need for temporal or step-wise treatment
\citep{wang2026tcod,zhong2026sod}. Our diagnosis asks a
complementary question: \textit{How is the reverse-KL supervision itself distributed across turns?}
We begin with the least processed measurement: the mean reverse-KL signal at each model turn during vanilla OPD.

\textbf{Setup.}
We use ALFWorld~\citep{shridhar2021alfworld} and Multi-Hop Search as example tasks.
Both diagnostics use vanilla OPD over 100 optimizer steps: ALFWorld with a Qwen3-4B~\citep{yang2025qwen3} student distilled from a Qwen3-8B-GRPO teacher (ALFWorld-4B), and Multi-Hop Search with a Qwen3.5-2B~\citep{yang2025qwen3} student from a Qwen3.5-9B-GRPO teacher (Multi-Hop Search-2B).
Turn-level means are computed only over trajectories that reach each turn. We plot the longest prefix where each turn is supported by at least 8 surviving trajectories, covers at least 10\% of turn-0 cases, and meets these criteria in at least 60\% of training steps. This flexible rule allows deeper analysis than a strict cutoff while avoiding noisy tails.
Dataset, environment, and teacher model details are in Appendix~\ref{app:envs}.

\begin{figure}[tbp]
\centering
\includegraphics[width=0.95\linewidth]{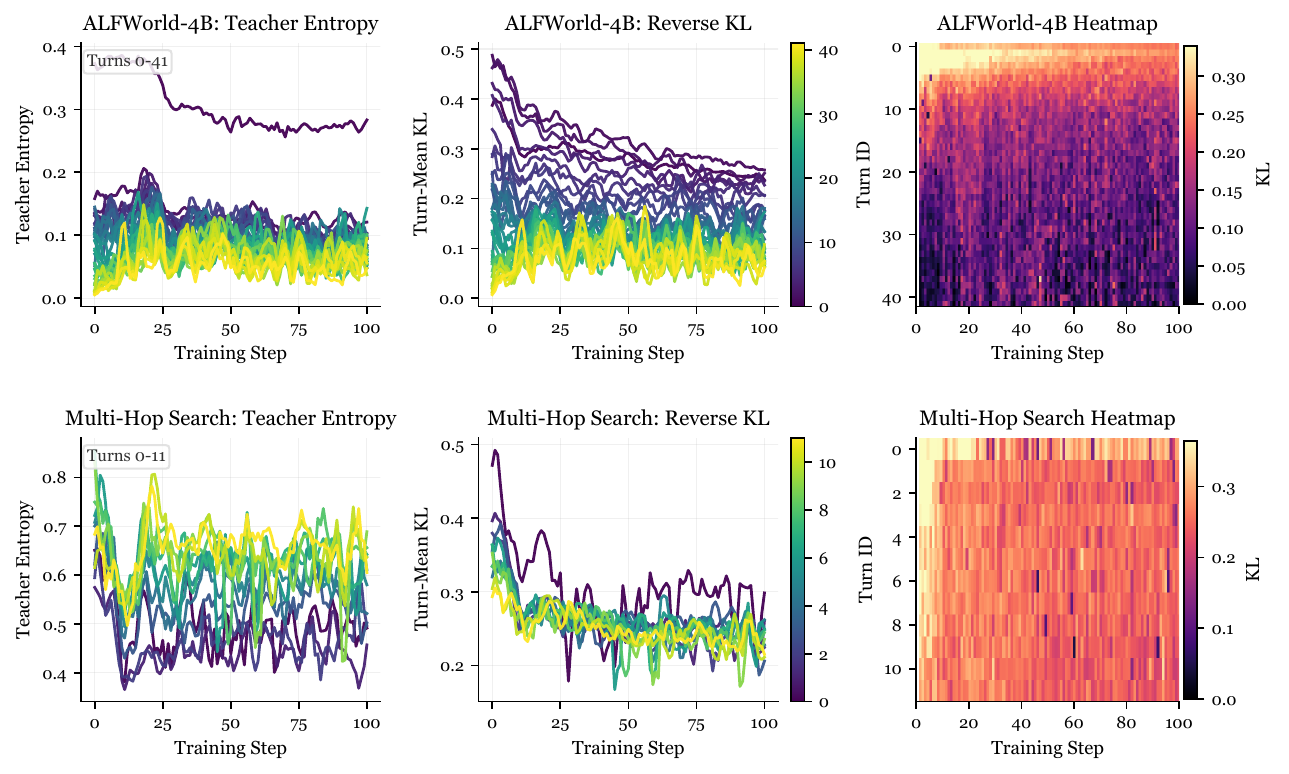}
\caption{Turn-resolved teacher uncertainty and reverse-KL under vanilla OPD.
The ALFWorld row uses the Qwen3-4B student baseline, and the Multi-Hop Search
row uses the Qwen3.5-2B student baseline. For each task, the left panel shows
smoothed per-turn teacher entropy, the middle panel shows smoothed per-turn
reverse-KL, and the right panel shows the raw turn-by-step reverse-KL heatmap.
All three panels use the same survivor-supported reliable turn prefix.}
\label{fig:diag-vanilla-turn-kl}
\vspace{-0.2cm}
\end{figure}

\textbf{Result.}
Figure~\ref{fig:diag-vanilla-turn-kl} plots the mean reverse-KL signal at each model turn.
Two patterns are visible. First, teacher uncertainty is itself turn-dependent
and task-dependent. In ALFWorld, deeper turns often have lower teacher
entropy. Multi-Hop Search shows a different entropy
profile: later turns can have higher teacher entropy, yet the
reverse-KL curves are still front-loaded.
Second, the reverse-KL signal is non-uniform and non-stationary over training.
On ALFWorld, early turns start with much larger KL and then drop quickly
during the first 20--30 optimizer steps, while a long tail of later turns
remains lower and noisier. Multi-Hop Search shows a shorter but still structured
profile: the first few turns have the largest KL, and the rest
settle into a narrower band after the initial alignment period. The immediate
conclusion is that vanilla agent OPD cannot be diagnosed by a single average KL.
However, the per-turn KL trend itself is ambiguous: late-turn decay is consistent both
with genuine alignment and with the teacher signal losing discriminative power.
We next test the latter directly.

\subsection{Outcome Separation Degrades and Inverts at Depth}
In a multi-turn setting, raw KL is not necessarily a reliable proxy for the
remaining student--teacher policy gap. We therefore ask a more outcome-oriented
question: at each turn, can KL distinguish rollouts that eventually succeed from
those that eventually fail, and how does this distinction evolve over turn indices and training phases?
For this purpose, define
\begin{equation}
G_t = K_t^{\mathrm{fail}} - K_t^{\mathrm{succ}},
\end{equation}
where $K_t^{\mathrm{fail}}$ and $K_t^{\mathrm{succ}}$ are the per-turn
reverse-KL means computed separately over failed and successful rollouts.
If raw KL were a clean local mismatch signal, failed rollouts should generally
show larger teacher corrections than successful rollouts at the same turn, leading to positive $G_t$.

\begin{figure}[htbp]
\centering
\includegraphics[width=0.85\linewidth]{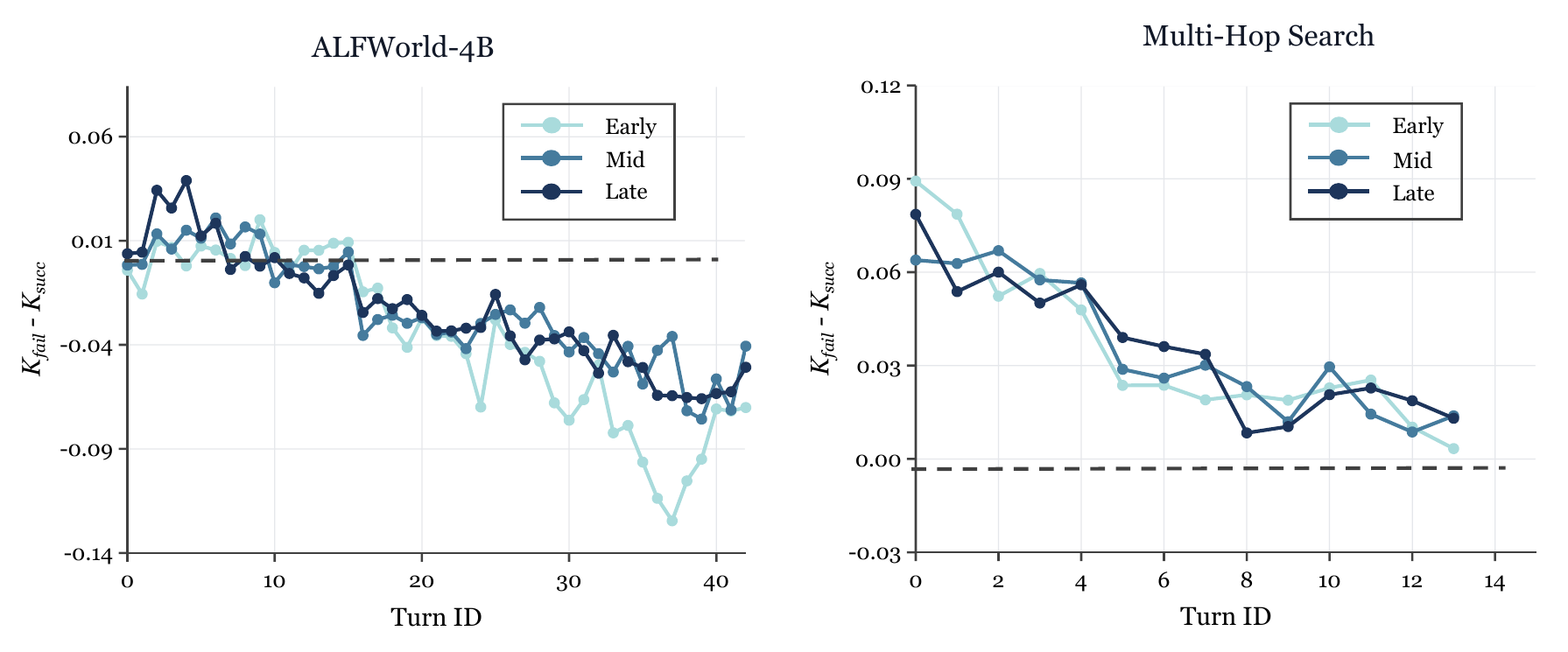}
\caption{Outcome separation of reverse-KL by turn under vanilla OPD. Each curve
plots the phase mean of $G_t=K_t^{\mathrm{fail}}-K_t^{\mathrm{succ}}$ over early
($0\le$ step $<30$), mid ($30\le$ step $<60$), or late ($60\le$ step $\le100$)
training. Positive values indicate larger KL on failed trajectories.}
\label{fig:diag-success-failure-gap}
\end{figure}

Figure~\ref{fig:diag-success-failure-gap} evaluates whether local KL distinguishes successful from failed trajectories. The sign of $G_t$ varies by task, but both show that deep-turn KL lacks outcome-predictive power.
For ALFWorld, $G_t$ is mostly negative and decreases with depth: successful trajectories often have higher per-turn KL than failed ones, especially at later turns. This is counterintuitive if KL reflects local correction demand, as failed rollouts should need more correction. We attribute this to branch-dependent compression: failed rollouts quickly fall into loops or template-like patterns where both models find the next-token prediction trivial, reducing KL artificially. In contrast, successful rollouts visit more diverse states where the teacher's corrections are less compressed.
In Multi-Hop Search, the gap has the expected positive sign: failed trajectories have larger KL. However, this separation is mostly present in early turns. In later turns, the gap shrinks, demonstrating that KL’s discriminative power is concentrated at the front and weakens with depth.

\subsection{A Contamination-Compression Mechanism}
\label{sec:diag:contamination}

The two measurements above leave a specific ambiguity. Raw reverse-KL decreases
along the reliable turn prefix, and its outcome-separation power also weakens
with depth and can even invert. Therefore a small late-turn KL can have two
different explanations: the student may have learned the teacher, or the
student-generated context may make both models follow the same local
continuation even when their policy-level preferences still differ. We formalize
this ambiguity with a contamination-compression model.

\textbf{Setup.}
Consider a supervised token position in model turn $t$, and let $c$ denote the
full on-policy context before that token, including previous student outputs and
environment/tool observations. Let $\pi_S(\cdot\mid c)$ and
$\pi_T(\cdot\mid c)$ be the next-token distributions of the student and teacher
under the same context. The observable raw KL at turn $t$ is
\begin{equation}
K_t
  = \mathbb{E}_{c\sim\mathcal{C}_t}
  \left[
  \KL\!\left(\pi_S(\cdot\mid c)\,\|\,\pi_T(\cdot\mid c)\right)
  \right],
\label{eq:observed-turn-kl}
\end{equation}
where $\mathcal{C}_t$ is the empirical distribution of supervised token contexts
collected at turn $t$. 

For a fixed context $c$, suppose part of the next-token distribution is nearly
forced by the context itself: repeated strings, formatting constraints, copied
entities, closing delimiters, or low-level continuations that both pretrained
language models assign high probability to. Let $p_F(\cdot\mid c)$ denote this
shared forced component. The remaining mass contains the free component, where
the student and teacher may genuinely disagree. We write
\begin{equation}
\pi_S(\cdot\mid c)
  = \lambda(c) p_F(\cdot\mid c)
    + (1-\lambda(c))p_S^{\mathrm{free}}(\cdot\mid c),
\qquad
\pi_T(\cdot\mid c)
  = \lambda(c) p_F(\cdot\mid c)
    + (1-\lambda(c))p_T^{\mathrm{free}}(\cdot\mid c),
\label{eq:forced-free-mixture}
\end{equation}
where $\lambda(c)\in[0,1]$ is the context-forced mass,
$p_S^{\mathrm{free}}$ is the student's free-component distribution, and
$p_T^{\mathrm{free}}$ is the teacher's free-component distribution. Define the
latent free-component disagreement as
\begin{equation}
\Delta_{\mathrm{free}}(c)
  =
  \KL\!\left(
  p_S^{\mathrm{free}}(\cdot\mid c)
  \,\|\,
  p_T^{\mathrm{free}}(\cdot\mid c)
  \right).
\label{eq:free-component-disagreement}
\end{equation}
Equation~\ref{eq:forced-free-mixture} assumes only that the student and teacher share the context-induced forced component; their free components may differ. In long-horizon settings, increasing student-generated context makes a larger portion of the next-token distribution surface-determined, compressing real decision differences into the remaining free component.

\begin{proposition}[Contamination compression]
\label{prop:contamination-compression}
Under the decomposition in Equation~\ref{eq:forced-free-mixture},
\begin{equation}
\KL\!\left(\pi_S(\cdot\mid c)\,\|\,\pi_T(\cdot\mid c)\right)
\le
(1-\lambda(c))\,
\KL\!\left(
p_S^{\mathrm{free}}(\cdot\mid c)\,\|\,
p_T^{\mathrm{free}}(\cdot\mid c)
\right).
\label{eq:contamination-bound}
\end{equation}
\end{proposition}

A full proof is provided in
Appendix~\ref{app:contamination-proof}.

Proposition~\ref{prop:contamination-compression} shows why raw KL is not
identifiable as pure policy disagreement. If
$\Delta_{\mathrm{free}}(c)>0$, the observable fraction of free disagreement
captured by raw KL satisfies
\begin{equation}
\rho(c)
  =
  \frac{
  \KL\!\left(\pi_S(\cdot\mid c)\,\|\,\pi_T(\cdot\mid c)\right)}
  {\Delta_{\mathrm{free}}(c)}
  \le 1-\lambda(c).
\label{eq:compression-ratio}
\end{equation}
Thus a high forced mass $\lambda(c)$ can compress the measured KL even when the
free-component disagreement remains nonzero.

\textbf{Why depth can increase compression.}
In a student rollout, the context at turn $t$ contains all earlier student
outputs and tool observations. Autoregressive generation is positively
self-conditioning: previous high-probability continuations make some future
continuations more predictable, and repeated or format-locked context can
increase the forced mass. It provides a structural explanation for the envelope
seen in Figure~\ref{fig:diag-vanilla-turn-kl}: as the rollout becomes longer,
the measured KL can decay because $(1-\lambda(c_t))$ contracts, not necessarily
because the free-component disagreement has vanished.

\textbf{Asymmetric compression and gap inversion.}
The success/failure gap reveals a stronger version of this mechanism: the
context-forced mass need not be symmetric across outcome branches. Failed
trajectories can exhibit stereotyped patterns, such as repeated invalid
actions, loops over the same subgoal, or template-like continuations. These
patterns are exactly the low-level linguistic regularity and local
self-correlation captured by the forced component $p_F$. Consequently, when the
teacher is conditioned on a failed context, its next-token distribution can be
more strongly locked into the local continuation, making the average forced mass
systematically larger on failed rollouts than on successful ones:
$\lambda_t^{\mathrm{fail}}>\lambda_t^{\mathrm{succ}}$. Successful trajectories,
in contrast, continue to traverse task states and observations; their contexts
remain more information-rich, so the teacher must still make genuine semantic
decisions and the forced mass is lower.

For $y\in\{\mathrm{succ},\mathrm{fail}\}$, let $\bar{\lambda}_t^{y}$ be the
average forced mass and $K_t^{y,\mathrm{free}}$ be the average free-component
reverse-KL among turn-$t$ contexts in outcome group $y$. Reading the compression
bound as a first-order approximation gives
$K_t^{y}\approx(1-\bar{\lambda}_t^{y})K_t^{y,\mathrm{free}}$, and therefore
\begin{equation}
G_t
  \approx
  (1-\bar{\lambda}_t^{\mathrm{fail}})K_t^{\mathrm{fail},\mathrm{free}}
  -
  (1-\bar{\lambda}_t^{\mathrm{succ}})K_t^{\mathrm{succ},\mathrm{free}} .
\label{eq:gap-compression}
\end{equation}
Equation~\ref{eq:gap-compression} explains why the gap can become negative. If
$\bar{\lambda}_t^{\mathrm{fail}}>\bar{\lambda}_t^{\mathrm{succ}}$, the failed
term is suppressed more strongly. Therefore $G_t$ can fall below zero even if
the uncompressed disagreement on failed rollouts is at least as large as that on
successful rollouts. This also explains why the curve can keep decreasing after
it has already become negative. As $t$ grows, failed trajectories accumulate
more degenerate context, so $\bar{\lambda}_t^{\mathrm{fail}}$ can keep
increasing; successful trajectories remain more task-driven, so
$\bar{\lambda}_t^{\mathrm{succ}}$ grows more slowly. The differential
$\bar{\lambda}_t^{\mathrm{fail}}-\bar{\lambda}_t^{\mathrm{succ}}$ therefore
widens with depth, producing the monotone downward trend observed in
Figure~\ref{fig:diag-success-failure-gap}.

\subsection{Loss Aggregation Creates Shallow Budget Concentration}
\label{sec:diag:turnnorm}

The diagnostics above show \textbf{where} local correction signal appears along
the turn axis. They do not show \textbf{where the optimizer spends its budget}.
This distinction matters because vanilla OPD updates the student through raw
reverse-KL. We therefore audit the realized KL loss mass assigned to each turn.

\begin{figure*}[htbp]
  \centering
  \includegraphics[width=\textwidth]{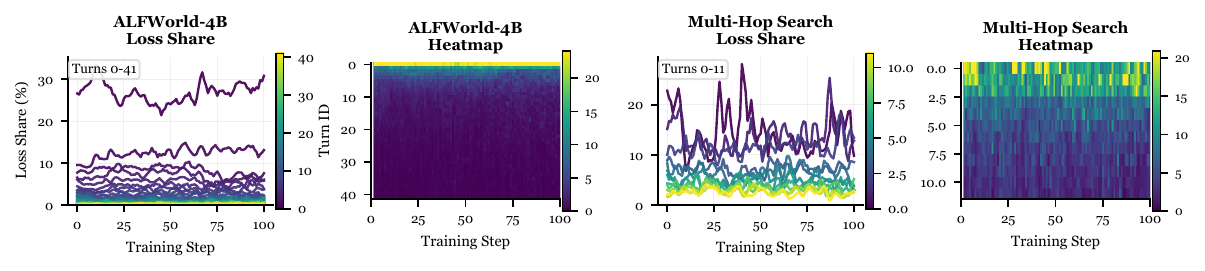}
  \caption[Turn-level KL loss-share under trajectory-level aggregation]{Turn-level KL loss-share under the vanilla trajectory-level KL objective. Line panels show one smoothed training-step curve per reliably observed turn; heatmaps show the raw step-by-turn loss-share values. The metric is the fraction of the current batch's raw KL loss mass contributed by turn $t$.}
  \label{fig:diag-loss-share}
\end{figure*}

Let a trajectory contain $T$ supervised model turns, and let turn $t$ contain
$n_t$ supervised tokens with token-level reverse-KL losses $\ell_{t,i}$. A
standard trajectory-level reduction is
\begin{equation}
L_{\mathrm{traj}}
  = \frac{1}{N}\sum_{t=1}^{T}\sum_{i=1}^{n_t}\ell_{t,i},
\qquad
N=\sum_{t=1}^{T}n_t .
\end{equation}
This reduction gives every supervised token the same base weight. Its token
share at turn $t$ is
\begin{equation}
b_t^{\mathrm{traj}} = \frac{n_t}{N}.
\end{equation}
The realized share of the batch KL loss mass is instead
\begin{equation}
s_t^{\mathrm{traj}}
  = \frac{\sum_{i=1}^{n_t}\ell_{t,i}}
         {\sum_{u=1}^{T}\sum_{j=1}^{n_u}\ell_{u,j}} .
\end{equation}
With this metric, we can now answer: \textbf{under the trajectory-level KL objective, what proportion of the current batch's KL loss mass is contributed by turn $t$?} We plot the turn-level KL loss-share for the ALFWorld and Multi-Hop Search tasks in Figure~\ref{fig:diag-loss-share}.

Figure~\ref{fig:diag-loss-share} shows that the KL loss is mainly concentrated in the early turns. In ALFWorld-4B, turn~0 alone uses about a quarter of the KL budget. The first three turns take nearly half. The reliable deep third receives only $3.6$--$4.5\%$ of the KL loss. In Multi-Hop Search, the effect is weaker, but the first three turns still take $38$--$40\%$ of the budget. The reliable deep third gets only $11$--$13\%$. This makes sense: not every rollout reaches the deep turns, so there are fewer tokens from deep turns in a batch. The KL loss in shallow turns is also larger. As a result, the KL loss is mostly assigned to earlier turns. The observations lead to two main conclusions:

\textbf{(1) Trajectory-level normalization creates a budget mismatch.}
This is not a claim that every deep turn is more valuable. The point is simpler:
the trajectory-level reducer spends loss by token mass and realized KL mass. It
can therefore keep most updates near the shallow prefix even when later turns
still contain correction signal. Table~\ref{tab:diag-loss-aggregation} quantifies
this with the same runs and the same reliable prefixes as
Figure~\ref{fig:diag-loss-share}.

\textbf{(2) A hard turn-level replacement may also be too crude.}
The other extreme gives each observed turn the same total weight. This removes
the shallow token-count bias, but it may also ignore reliability. In ALFWorld-4B,
the reliable deep third has only $31$--$42\%$ of the shallow prefix's raw KL. In
Multi-Hop Search, the deep third has comparable raw KL, but only about
$16$--$17\%$ survivor support. Equalizing all turns from the beginning would
therefore amplify weaker or lower-support estimates too early.

\begin{table*}[tbp]
  \centering
  \caption[Loss-allocation diagnostics]{Loss-allocation diagnostics from vanilla OPD baselines.
  Deep/shallow ratios compare the deepest third of the reliable turn prefix with
  the shallowest third, using the same prefixes as
  Figure~\ref{fig:diag-loss-share}. Raw KL is the actual optimization signal.
  ``Deep support'' is survivor coverage in the deepest third, and ``Deep loss
  budget'' is the fraction of total raw-KL loss mass assigned to that deepest
  third by the trajectory-level objective.}
  \label{tab:diag-loss-aggregation}
  \small
  \setlength{\tabcolsep}{6pt}
  \begin{tabular}{@{}>{\raggedright\arraybackslash}m{3cm}>{\centering\arraybackslash}m{1.4cm}>{\centering\arraybackslash}m{2.3cm}>{\centering\arraybackslash}m{2.2cm}>{\centering\arraybackslash}m{2.2cm}@{}}
  \toprule
  Task & Phase & \shortstack{Deep/Shallow\\raw KL} & \shortstack{Deep\\support} & \shortstack{Deep loss\\budget} \\
  \midrule
  ALFWorld & early & 31\% & 23.0\% & 3.6\% \\
  ALFWorld & late & 42\% & 18.2\% & 4.5\% \\
  Multi-Hop Search & early & 90\% & 17.3\% & 12.9\% \\
  Multi-Hop Search & late & 92\% & 15.5\% & 11.1\% \\
  \bottomrule
  \end{tabular}
\end{table*}
\FloatBarrier


\subsection{Summary: Two Budget Mismatches}
\label{sec:diag:fail}

The diagnosis isolates two budget mismatches in vanilla agent OPD:

\textbf{First, an external mismatch.} Rollout depth is fixed, but the
survivor-weighted local correction signal and the survivor population both vary
sharply with turn depth. Collecting to the maximum horizon can spend rollout
compute on turns whose observable correction mass is low or whose estimates are
dominated by a small survivor subset. This motivates an adaptive rollout-depth
controller that decides \textbf{how far to collect}.

\textbf{Second, an internal mismatch.} After a trajectory is collected,
trajectory-level normalization starts from uniform token weighting and realizes a
shallow KL-mass concentration. 
However, in agent tasks, turns with few tokens are not necessarily less important. After mastering initial interactions, remaining errors are often in deeper decision turns, yet the token-weighted objective assigns little loss budget to these steps.
This motivates a progressive turn-level weighting
controller that decides \textbf{where the collected loss mass should go}.

These two failures have the same underlying structure: the useful unit of
supervision in a long-horizon agent is not a flat token position, but a
turn-conditioned decision embedded in an evolving interaction trace.



%% file: sections/method.tex
\section{Method: TurnOPD}
\label{sec:method}
Based on the analysis in the diagnosis section, we identify two main issues in applying OPD to agents. First, the per-turn KL signal decreases sharply with turn depth, especially in the early training stages. The KL loss is dominated by shallow turns. This suggests that it is not necessary to wait until the end of the full rollout to compute the KL loss. Shortening the rollout horizon can improve overall training efficiency. Second, after a trajectory is collected, trajectory-level normalization assigns loss mass according to token count. This approach can neglect deeper decision turns once the shallow interaction patterns have converged. 

In this section, we propose \textbf{TurnOPD}, a turn-level budgeting strategy for efficient OPD of long-horizon agents. TurnOPD consists of two budget controllers. The first controller adaptively budgets rollout depth. The second controller progressively shifts KL loss allocation from trajectory-level matching toward turn-balanced decision refinement.

\subsection{External Mismatch: Adaptive Rollout-Depth Budgeting}
\label{sec:method:A}
A key limitation identified in the diagnosis is the mismatch between rollout depth and the effective utility of supervision: the optimal trajectory length for data collection should account for both task and student progress to maximize the value of supervision per unit cost. Let $H^\star$ denote this optimal (but unobservable) rollout horizon, the depth yielding the greatest marginal validation improvement for the supervision cost. The existence of $H^\star$ is guaranteed by an efficiency–coverage tradeoff: too shallow a rollout under-explores, while overly deep rollouts waste computation on low-signal regions. A full proof is provided in Appendix~\ref{app:rollout-depth-existence}.

However, $H^\star$ cannot be measured directly during training. Instead, TurnOPD estimates it online by synthesizing two complementary signals: an efficiency-centric proxy and a coverage-based lower bound. Specifically, we define the rollout-depth budget controller as
\begin{equation}
H_{\mathrm{ctrl}} = \max\big(\Heff,\, \Hcov\big),
\end{equation}
where $\Heff$ captures the centroid of effective supervision and $\Hcov$ enforces minimum task completion coverage.

\textbf{Efficiency-side Mass.} To construct $\Heff$, we treat the reverse-KL signal over turns as a survivor-weighted distribution of distillation mass. Raw KL may be unreliable as a behavioral-error measure, but still useful as an observable supervision-value and cost proxy. For each turn $t$, let $K_t$ be the mean per-token reverse-KL on probed student rollouts and $n_t/n_0$ the on-policy survivor probability. We define
\begin{equation}
m_t = [K_t]_+\,\cdot \frac{n_t}{n_0}, \qquad
q_t = \frac{m_t}{\sum_j m_j + \epsilon},
\end{equation}
where $[K_t]_+$ denotes the positive part of $K_t$\footnote{%
In practice, $K_t$ can be slightly negative due to noise, so $[K_t]_+$ ensures $m_t$ is stable and non-negative.
}.
$m_t$ estimates the expected value of distillation signal contributed by turn $t$, and $q_t$ normalizes over all turns. The centroid and its discretized projection are then
\begin{equation}
\bar H_{\mathrm{eff}} = \sum_t t\,q_t, \qquad
\Heff = \mathrm{round}(\bar H_{\mathrm{eff}}).
\end{equation}
This first-moment statistic responds adaptively: when meaningful supervision mass is concentrated in shallow turns, $\Heff$ remains shallow; when a correction tail persists at deeper turns, $\Heff$ deepens accordingly. Relative weighting ensures robustness to variation in absolute KL magnitude.

\textbf{Coverage-side Lower Bound.} To prevent overly aggressive truncation, the coverage component sets a minimum rollout depth based on successful completions:
\begin{equation}
\Hcov = \hat Q_p(L_{\mathrm{succ}}) = \min\{H: F_{\mathrm{succ}}(H)\ge p\}, \qquad
F_{\mathrm{succ}}(H) = 1 - \frac{n^{\mathrm{succ}}_{H+1}}{n^{\mathrm{succ}}_1}, \quad p=0.80,
\end{equation}
where $\Hcov$ is the $p$-quantile of the success-conditioned completion depth. This ensures the rollout is deep enough to include at least 80\% of successful trajectories, as estimated from periodic probe rollouts.

\textbf{Dynamic Update.}
The controller updates $\Heff$ on every batch and $\Hcov$ only on periodic full-length probes; routine steps use only the capped depth $\hat H_k$. Since capped batches cannot reveal turns beyond the cap, $\Heff$ computed on them cannot raise the depth, so growth relies on the periodic full-depth probes.

After every step, TurnOPD computes $H_{\mathrm{ctrl},k} = \max(H_{\mathrm{eff},k}, H_{\mathrm{cov},k})$ and updates rollout depth with exponential smoothing:
\begin{equation}
\bar H_k = (1-\alpha_{\mathrm{ema}})\bar H_{k-1}
          + \alpha_{\mathrm{ema}} H_{\mathrm{ctrl},k}, \qquad
\hat H_{k+1} = \mathrm{clip}\big(\mathrm{round}(\bar H_k)+1,\,
H_{\min},\,H_{\max}\big).
\end{equation}
The $+1$ compensates for zero-based indexing.

\subsection{Internal Mismatch: Progressive Turn-Normalized Loss Budgeting}
\label{sec:method:B}

A second challenge is the internal budget mismatch: with standard trajectory-level normalization, most of the KL loss is assigned to shallow turns, so deeper turns often receive little or no supervision—especially at later training stages, when optimizing long-horizon decisions becomes crucial. This is reasonable early on (when shallow errors dominate), but later hampers correction of harder, deeper decisions.

TurnOPD remedies this by introducing a linear blend between trajectory-level and turn-level normalization, shifting the loss allocation over the course of training. Specifically, let $n_t$ be the number of tokens at turn $t$ (with $T$ total turns). The standard trajectory-based weight for turn $t$ is $q_t^{\mathrm{traj}}=n_t/\sum_j n_j$, while the uniform turn-wise weight is $q_t^{\mathrm{turn}}=1/T$. The final weight is a linear interpolation:
\[
q_t^{\mathrm{blend}} = (1-\alpha)\,q_t^{\mathrm{traj}} + \alpha\,q_t^{\mathrm{turn}},
\]
where the blend coefficient $\alpha$ is tied to normalized training progress, e.g., $\alpha=\mathrm{clip}((\text{progress}-s)/(e-s),\,0,\,1)$ with $\text{progress} = k/K$ (step $k$ of $K$, and typically $s=0, e=1$).

Initially ($\alpha=0$), loss follows token mass, ensuring stability in early stages. As $\alpha$ grows, supervision smoothly shifts to cover all turns more equally, explicitly mitigating turn-depth imbalance and enabling better learning of deep decisions.

\subsection{Unification and Algorithm}
\label{sec:method:alg}

The two interventions regulate complementary resources: rollout depth determines how much interaction is collected, while loss normalization dictates how supervision is allocated over the collected tokens. For the complete training algorithm and all hyperparameters, see Appendix~\ref{app:hyper}.

%% file: sections/experiments.tex
\section{Experiments}
\label{sec:exp}
We evaluate TurnOPD on three representative agent tasks: ALFWorld (embodied text planning), Multi-Hop Search, and WebShop (web navigation), using three different student-teacher pairs that span multiple model sizes and model families. 
We test whether TurnOPD achieves a better accuracy--time frontier across different environments and models.

\textbf{Students and teachers.} 
Students are distilled from stronger, task-specialized GRPO-trained teachers: ALFWorld uses Qwen3-1.7B/4B students with a Qwen3-8B-GRPO teacher; WebShop uses Qwen3-1.7B and Qwen3-8B-GRPO; Multi-Hop Search uses Qwen3.5-2B and Qwen3.5-9B-GRPO. All methods use the same OPD reverse-KL training stack.

\textbf{Baselines and Metrics.} We evaluate three primary methods: vanilla OPD, TCOD-F2B~\citep{wang2026tcod}, and TurnOPD. To ensure a fair and comprehensive comparison, we conduct evaluations under two regimes: \textit{Least-Time}—where we compare all methods based on accuracies achieved at the minimal wall-clock time required for any method to reach 100 training steps; and \textit{Same-Step}—where each method is evaluated after exactly 100 training steps. Accuracy for each method is reported as the average accuracy over the last four evaluation points (each point is calculated using Avg@4) before the corresponding cutoff. We also report the standard deviation of the accuracy over the last four evaluation points.
In addition, we include the performance of the GRPO-trained teacher models as reference points. Note that teacher accuracy is intended for context and not as a baseline for efficiency comparisons.

\providecommand{\avgstd}[1]{{\normalfont\tiny$\pm$#1}}

\begin{table*}[tbp]
    \centering
    \caption{Main comparison across tasks. Overall accuracy is reported as Avg@4
    before the corresponding cutoff; the smaller $\pm$ term reports the
    standard deviation across the same four evaluation checkpoints. Wall time
    is the cumulative per-step
    training time through 100 optimizer steps, and speedup is relative to
    vanilla OPD within each task--student group. \textbf{Bold} numbers mark the best
    student-training method.}
    \label{tab:main}
    \scriptsize
    \setlength{\tabcolsep}{5pt}
    \begin{tabular}{>{\raggedright\arraybackslash}m{1.8cm}
        >{\centering\arraybackslash}m{1.6cm}
        >{\centering\arraybackslash}m{2.6cm}
        >{\raggedright\arraybackslash}m{1.5cm}
        >{\centering\arraybackslash}m{1.7cm}
        >{\centering\arraybackslash}m{1.7cm}
        >{\centering\arraybackslash}m{1.2cm}
        >{\centering\arraybackslash}m{1.3cm}}
    \toprule
    \textbf{Task}
    & \textbf{Student}
    & \textbf{Teacher}
    & \textbf{Method}
    & \makecell[c]{\textbf{Overall Avg@4}\\\textbf{Least-Time}}
    & \makecell[c]{\textbf{Overall Avg@4}\\\textbf{Same-Step}}
    & \makecell[c]{\textbf{Same Step}\\\textbf{Wall (h)}}
    & \makecell[c]{\textbf{Same Step}\\\textbf{Speedup}} \\
    \midrule
     \multirow{5}{1.95cm}{\raggedright ALFWorld} & \multirow{5}{1.85cm}{\centering Qwen3-1.7B} & \multirow{5}{2.45cm}{\centering Qwen3-8B-GRPO} & Zero-Shot & 0.00 & 0.00 & -- & -- \\
     &  &  & Teacher & 90.75 & 90.75 & -- & -- \\
     &  &  & Vanilla OPD & 73.52\avgstd{1.84} & 83.00\avgstd{1.69} & 4.42 & 1.00$\times$ \\
     &  &  & TCOD-F2B & 80.06\avgstd{1.43} & 80.06\avgstd{1.43} & \textbf{1.87} & \textbf{2.37$\times$} \\
     &  &  & \cellcolor{blue!9}\textbf{TurnOPD} & \cellcolor{blue!9}\textbf{85.60}\avgstd{0.95} & \cellcolor{blue!9}\textbf{86.29}\avgstd{0.48} & \cellcolor{blue!9}1.93 & \cellcolor{blue!9}2.29$\times$ \\
    \midrule
     \multirow{5}{1.95cm}{\raggedright ALFWorld} & \multirow{5}{1.85cm}{\centering Qwen3-4B} & \multirow{5}{2.45cm}{\centering Qwen3-8B-GRPO} & Zero-Shot & 8.17 & 8.17 & -- & -- \\
     &  &  & Teacher & 90.75 & 90.75 & -- & -- \\
     &  &  & Vanilla OPD & 90.79\avgstd{0.61} & 91.81\avgstd{0.83} & 2.86 & 1.00$\times$ \\
     &  &  & TCOD-F2B & 86.50\avgstd{1.88} & 86.50\avgstd{1.88} & \textbf{1.89} & \textbf{1.51$\times$} \\
     &  &  & \cellcolor{blue!9}\textbf{TurnOPD} & \cellcolor{blue!9}\textbf{91.73}\avgstd{1.41} & \cellcolor{blue!9}\textbf{92.21}\avgstd{0.42} & \cellcolor{blue!9}2.16 & \cellcolor{blue!9}1.33$\times$ \\
    \midrule
     \multirow{5}{1.95cm}{\raggedright Multi-Hop Search} & \multirow{5}{1.85cm}{\centering Qwen3.5-2B} & \multirow{5}{2.45cm}{\centering Qwen3.5-9B-GRPO} & Zero-Shot & 36.08 & 36.08 & -- & -- \\
     &  &  & Teacher & 59.00 & 59.00 & -- & -- \\
     &  &  & Vanilla OPD & 45.77\avgstd{0.48} & \textbf{47.82}\avgstd{0.92} & 4.45 & 1.00$\times$ \\
     &  &  & TCOD-F2B & 45.64\avgstd{1.06} & 47.77\avgstd{1.21} & 3.80 & 1.17$\times$ \\
     &  &  & \cellcolor{blue!9}\textbf{TurnOPD} & \cellcolor{blue!9}\textbf{47.24}\avgstd{1.03} & \cellcolor{blue!9}47.24\avgstd{1.03} & \cellcolor{blue!9}\textbf{2.94} & \cellcolor{blue!9}\textbf{1.51$\times$} \\
    \midrule
     \multirow{5}{1.95cm}{\raggedright WebShop} & \multirow{5}{1.85cm}{\centering Qwen3-1.7B} & \multirow{5}{2.45cm}{\centering Qwen3-8B-GRPO} & Zero-Shot & 25.80 & 25.80 & -- & -- \\
     &  &  & Teacher & 84.46 & 84.46 & -- & -- \\
     &  &  & Vanilla OPD & 76.98\avgstd{0.79} & 81.65\avgstd{1.47} & 1.57 & 1.00$\times$ \\
     &  &  & TCOD-F2B & 80.45\avgstd{1.57} & 81.66\avgstd{0.89} & 1.33 & 1.18$\times$ \\
     &  &  & \cellcolor{blue!9}\textbf{TurnOPD} & \cellcolor{blue!9}\textbf{82.80}\avgstd{0.75} & \cellcolor{blue!9}\textbf{82.80}\avgstd{0.75} & \cellcolor{blue!9}\textbf{1.24} & \cellcolor{blue!9}\textbf{1.26$\times$} \\
    \bottomrule
    \end{tabular}
    \vspace{-0.2cm}
\end{table*}

\subsection{Overall Performance}
\label{sec:exp:main}
\begin{figure}[tbp]
    \centering
    \includegraphics[width=\textwidth]{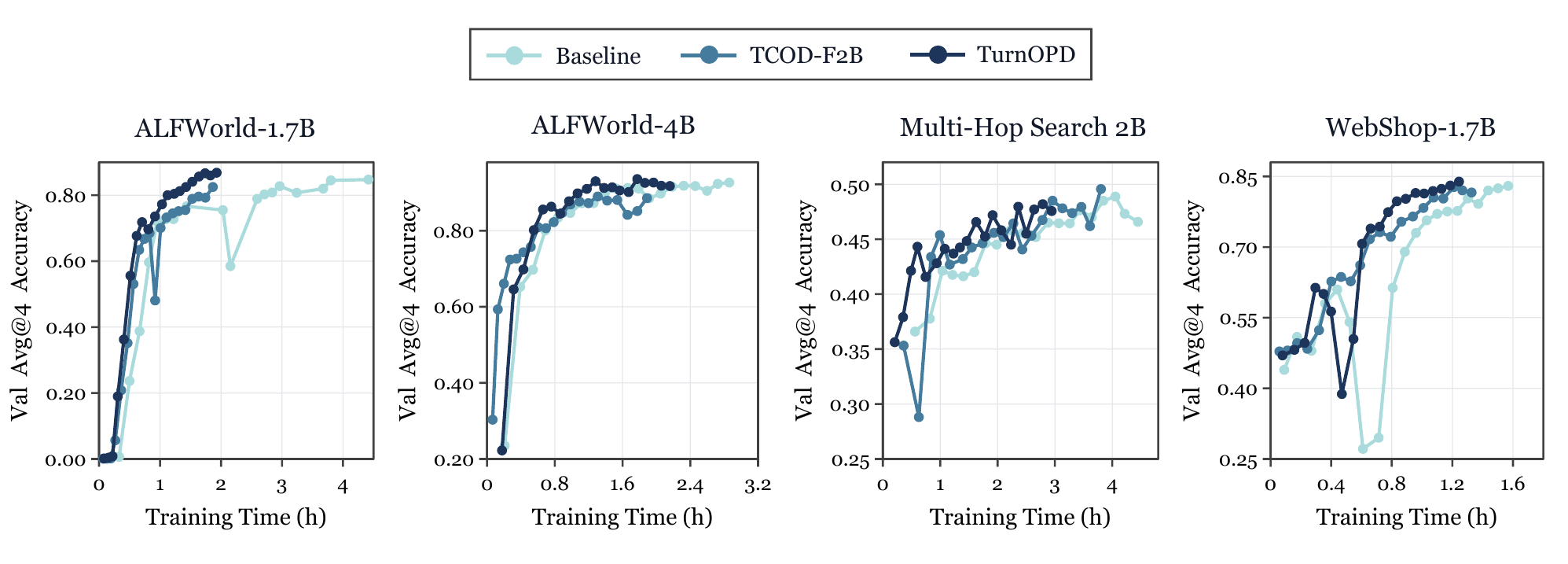}
    \caption{Main-result iso-time efficiency.
    The x-axis is cumulative training time, computed as the running sum of
    per training step time.
    TurnOPD improves the accuracy--time frontier on each task type.}
    \label{fig:crosstask}
    \end{figure}

Table~\ref{tab:main} and Figure~\ref{fig:crosstask} summarize the main results across three long-horizon tasks and multiple student-teacher pairs. Both vanilla OPD and TCOD-F2B serve as strong baselines, but TurnOPD consistently improves the overall accuracy--time tradeoff.

\textbf{Accuracy.} Under the Least-Time setting, TurnOPD achieves the highest overall avg@4 across all task and model combinations among the student-training methods. For the Same-Step setting, TurnOPD matches or surpasses the best baseline in most cases. For instance, on ALFWorld with a Qwen3-1.7B student, TurnOPD reaches 86.29 avg@4, compared to 83.00 for vanilla OPD and 80.06 for TCOD-F2B. On Multi-Hop Search, vanilla OPD keeps a small Same-Step advantage, but TurnOPD gives the best Least-Time accuracy, showing that it reaches stronger performance under the shared wall-clock budget. Tables~\ref{tab:alfworld-category} and~\ref{tab:multihop-category} further provide the task-specific breakdowns. On ALFWorld, TurnOPD is the strongest student-training method on nearly all categories for both student sizes. On Multi-Hop Search, TurnOPD improves PopQA and 2Wiki and gives the best overall Least-Time score.  Notably, with the Qwen3-4B student, TurnOPD even exceeds the ALFWorld teacher reference in overall avg@4.

\textbf{Training Efficiency and Wall-Clock Speedup.} Beyond accuracy, TurnOPD also reduces training time. On ALFWorld-1.7B, TurnOPD reduces the 100-step wall-clock time from 4.42 hours for vanilla OPD to 1.93 hours while improving accuracy. Similar improvements appear on Multi-Hop Search (2.94h vs. 4.45h) and WebShop (1.24h vs. 1.57h). Thus TurnOPD is not only more accurate under the common wall-clock budget, but also substantially accelerates vanilla OPD.

In summary, these comprehensive results establish TurnOPD as a strong performer on both accuracy and efficiency axes. Its turn-level budgeting approach yields robust accuracy improvements with less computation, making it attractive for both research and real-world applications where computational resources are a critical constraint.

\begin{table*}[tbp]
    \centering
    \caption{ALFWorld category-level accuracy under the Least-Time protocol.
    Category scores report avg@4 before the same wall-clock cutoff used in
    Table~\ref{tab:main}.}
    \label{tab:alfworld-category}
    \scriptsize
    \setlength{\tabcolsep}{5pt}
    \begin{tabular}{>{\raggedright\arraybackslash}m{1.6cm}
        >{\raggedright\arraybackslash}m{1.4cm}
        >{\centering\arraybackslash}m{1.3cm}
        >{\centering\arraybackslash}m{1.3cm}
        >{\centering\arraybackslash}m{1.3cm}
        >{\centering\arraybackslash}m{1.3cm}
        >{\centering\arraybackslash}m{1.3cm}
        >{\centering\arraybackslash}m{1.3cm}
        >{\centering\arraybackslash}m{2cm}}
    \toprule
    \textbf{Student}
    & \textbf{Method}
    & \textbf{2Obj}
    & \textbf{Heat}
    & \textbf{Cool}
    & \textbf{Clean}
    & \textbf{Place}
    & \textbf{Look}
    & \makecell[c]{\textbf{Overall Avg@4}\\\textbf{Least-Time}} \\
    \midrule
    \multirow{5}{1.85cm}{\raggedright Qwen3-1.7B} & Zero-Shot & 0.00 & 0.00 & 0.00 & 0.00 & 0.00 & 0.00 & 0.00 \\
     & Teacher & 92.50 & 93.40 & 84.09 & 92.34 & 91.49 & 91.28 & 90.75 \\
     & Vanilla OPD & 76.41\avgstd{2.77} & 73.82\avgstd{2.15} & 65.45\avgstd{3.51} & 73.39\avgstd{2.57} & 75.93\avgstd{2.27} & 78.34\avgstd{2.68} & 73.52\avgstd{1.84} \\
     & TCOD-F2B & 82.66\avgstd{2.70} & 80.19\avgstd{1.20} & 74.43\avgstd{3.63} & 79.03\avgstd{0.99} & 82.18\avgstd{2.14} & 83.87\avgstd{1.76} & 80.06\avgstd{1.43} \\
     & \cellcolor{blue!9}\textbf{TurnOPD} & \cellcolor{blue!9}\textbf{87.97}\avgstd{0.27} & \cellcolor{blue!9}\textbf{85.97}\avgstd{2.09} & \cellcolor{blue!9}\textbf{80.34}\avgstd{0.38} & \cellcolor{blue!9}\textbf{84.58}\avgstd{1.55} & \cellcolor{blue!9}\textbf{87.10}\avgstd{2.67} & \cellcolor{blue!9}\textbf{89.53}\avgstd{2.06} & \cellcolor{blue!9}\textbf{85.60}\avgstd{0.95} \\
    \midrule
    \multirow{5}{1.85cm}{\raggedright Qwen3-4B} & Zero-Shot & 4.38 & 5.19 & 3.64 & 5.24 & 11.70 & 21.51 & 8.17 \\
     & Teacher & 92.50 & 93.40 & 84.09 & 92.34 & 91.49 & 91.28 & 90.75 \\
     & Vanilla OPD & 92.97\avgstd{0.52} & 89.50\avgstd{0.39} & 90.34\avgstd{2.21} & 90.02\avgstd{1.08} & \textbf{91.49}\avgstd{1.95} & 91.28\avgstd{0.92} & 90.79\avgstd{0.61} \\
     & TCOD-F2B & 88.91\avgstd{1.20} & 85.26\avgstd{2.86} & 85.00\avgstd{2.51} & 84.17\avgstd{2.29} & 87.23\avgstd{3.24} & 90.26\avgstd{1.86} & 86.50\avgstd{1.88} \\
     & \cellcolor{blue!9}\textbf{TurnOPD} & \cellcolor{blue!9}\textbf{94.53}\avgstd{2.44} & \cellcolor{blue!9}\textbf{91.04}\avgstd{1.45} & \cellcolor{blue!9}\textbf{90.68}\avgstd{0.75} & \cellcolor{blue!9}\textbf{91.94}\avgstd{2.26} & \cellcolor{blue!9}90.96\avgstd{1.36} & \cellcolor{blue!9}\textbf{91.86}\avgstd{2.36} & \cellcolor{blue!9}\textbf{91.73}\avgstd{1.41} \\
    \bottomrule
    \end{tabular}
    \vspace{-0.2cm}
\end{table*}

\begin{table*}[tbp]
    \centering
    \caption{Multi-Hop Search category-level accuracy under the Least-Time
    protocol. Category scores report avg@4 over the four held-out QA sources.}
    \label{tab:multihop-category}
    \scriptsize
    \setlength{\tabcolsep}{5pt}
    \begin{tabular}{>{\raggedright\arraybackslash}m{2.3cm}
        >{\centering\arraybackslash}m{1.5cm}
        >{\centering\arraybackslash}m{1.6cm}
        >{\centering\arraybackslash}m{1.6cm}
        >{\centering\arraybackslash}m{1.7cm}
        >{\centering\arraybackslash}m{2.2cm}}
    \toprule
    \textbf{Method}
    & \textbf{PopQA}
    & \textbf{NQ}
    & \textbf{2Wiki}
    & \textbf{HotpotQA}
    & \makecell[c]{\textbf{Overall Avg@4}\\\textbf{Least-Time}} \\
    \midrule
    Zero-Shot & 38.25 & 29.00 & 39.25 & 37.81 & 36.08 \\
    Teacher & 59.50 & 52.50 & 72.00 & 52.00 & 59.00 \\
    Vanilla OPD & 48.00\avgstd{0.31} & \textbf{40.88}\avgstd{1.07} & 50.81\avgstd{1.15} & \textbf{43.39}\avgstd{0.53} & 45.77\avgstd{0.48} \\
    TCOD-F2B & 47.81\avgstd{1.01} & 39.38\avgstd{0.38} & 53.25\avgstd{3.05} & 42.12\avgstd{1.26} & 45.64\avgstd{1.06} \\
    \cellcolor{blue!9}\textbf{TurnOPD} & \cellcolor{blue!9}\textbf{49.38}\avgstd{1.64} & \cellcolor{blue!9}40.69\avgstd{1.12} & \cellcolor{blue!9}\textbf{56.00}\avgstd{2.49} & \cellcolor{blue!9}42.89\avgstd{0.98} & \cellcolor{blue!9}\textbf{47.24}\avgstd{1.03} \\
    \bottomrule
    \end{tabular}
    \vspace{-0.2cm}
\end{table*}

\subsection{Diagnostic--Controller Alignment}
\label{sec:exp:diag}
The main results confirm that TurnOPD effectively improves both the accuracy and compute efficiency frontiers. To further validate the effectiveness of the adaptive controller, we examine whether the controller’s behavior aligns with our diagnostic metrics that motivated its design.

\begin{figure}[htbp]
\centering
\includegraphics[width=0.9\linewidth]{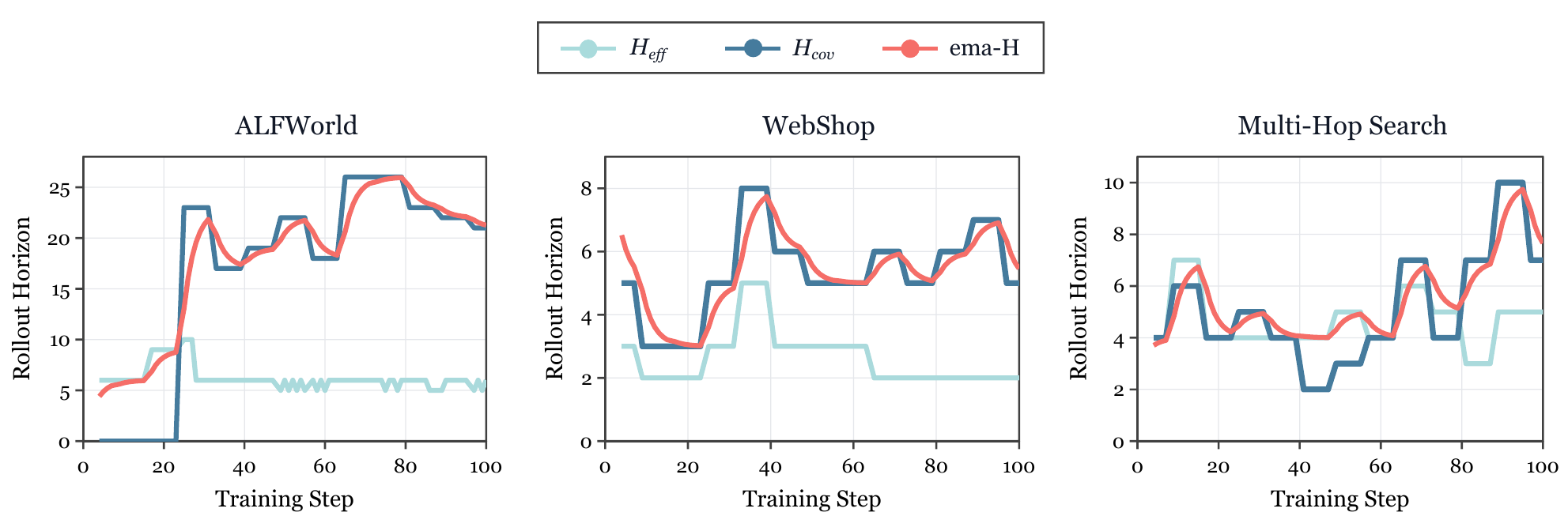}
\caption{Rollout-depth diagnostic replay. For each TurnOPD run, we plot the
survivor-weighted raw-KL centroid $\Heff$, the success-coverage lower bound
$\Hcov$, and the counterfactual EMA horizon $\hat H_{\mathrm{EMA}}$
obtained by replaying $\max(\Heff,\Hcov)$. Values use the 0-based turn-index
convention; the applied rollout cap is approximately
$\mathrm{round}(\hat H)+1$.}
\label{fig:diag-align}
\end{figure}

Figure~\ref{fig:diag-align} illustrates the dynamics of TurnOPD's rollout-depth adaptation. Specifically, we plot the survivor-weighted raw-KL centroid $\Heff$ (representing the efficiency arm) and the success-conditioned completion quantile $\Hcov$ (representing the coverage arm) as defined in Section~\ref{sec:method:A}. The controller adaptively targets the maximum of these two quantities at each step. During initial warm-up phases before sufficient successful rollouts appear, $\Hcov$ is set to zero to ensure that the observed KL centroid remains visible. Across tasks, we observe distinct controller behaviors: (1) in ALFWorld, the coverage constraint ($\Hcov$) quickly dominates once successful trajectories emerge, encouraging longer rollouts; (2) for WebShop, the controller maintains a moderate coverage lower bound throughout; and (3) in Multi-Hop Search, the balance shifts dynamically between the efficiency and coverage arms. These plots confirm the intended division of labor: $\Heff$ ensures nontrivial efficiency, while $\Hcov$ prevents the controller from selecting overly short rollouts that would limit coverage of successful completions.

%% file: sections/ablation.tex
\section{Ablation Studies and Analysis}
\label{sec:exp:abl}

The previous section demonstrates that the learned controller tracks the diagnostic signals underlying the design of TurnOPD. We now turn to a more fine-grained question: How do individual budget controllers in TurnOPD contribute to its overall performance? 

To answer this, we conduct ablation studies on ALFWorld-1.7B (Qwen3-1.7B distilled from Qwen3-8B-GRPO on ALFWorld task) using the same 100-step protocol as in the main experiment. The analysis proceeds in three parts. First, we disentangle the two core interventions to evaluate whether each is effective on its own. Second, we ablate the KL normalization scheme, comparing trajectory-level, strict turn-level, and linear turn-level reductions. Third, we analyze the effect of altering the coverage floor, which determines the conservativeness of the adaptive-depth controller.

\subsection{Intervention Component Decomposition}
\label{sec:exp:intervention-decomposition}

This subsection analyzes the individual contribution of each TurnOPD budget controller to overall performance.

\textbf{Setup.}
We conduct experiments on the ALFWorld task.
The \textit{Adaptive Depth} variant applies the adaptive rollout-depth budget controller while keeping the original trajectory-level KL reduction unchanged, to test if simply shortening rollout depth is sufficient. In contrast, the \textit{Linear blend norm} variant maintains the full-horizon OPD rollout depth but replaces the KL reduction with a linear turn-level normalizer, assessing whether rebalancing loss allocation alone can be effective. The full TurnOPD approach integrates both budget controllers. We report the overall Avg@4 accuracy under full training, averaging the last four evaluation checkpoints.

\begin{figure}[tbp]
    \centering
    \begin{minipage}[c]{0.45\textwidth}
        \centering
        \includegraphics[width=\linewidth]{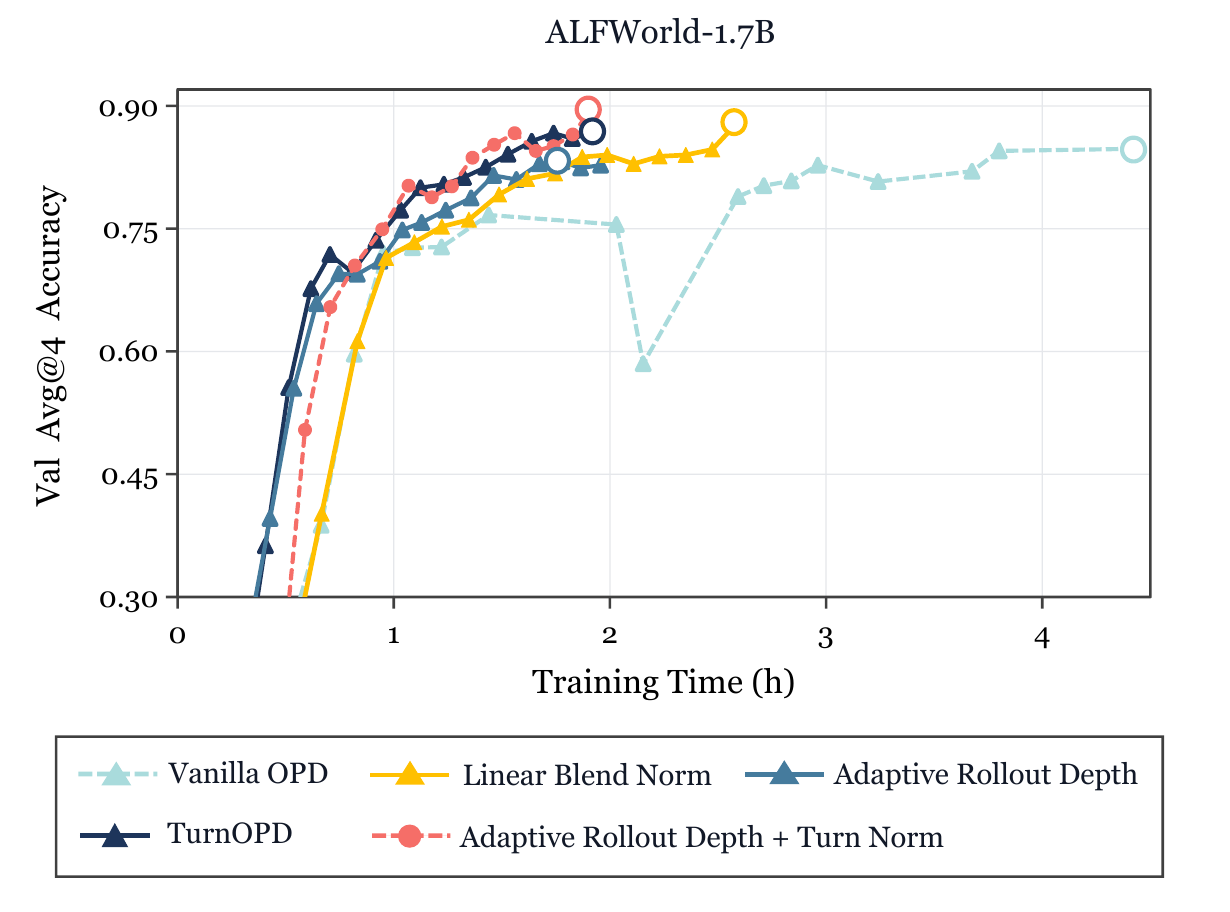}
        \captionsetup{hypcap=false}
        \captionof{figure}{ALFWorld-1.7B component ablations.}
        \label{fig:ablation}
    \end{minipage}\hfill
    \begin{minipage}[c]{0.53\textwidth}
        \centering
        \captionsetup{hypcap=false}
        \captionof{table}{Intervention decomposition on ALFWorld-1.7B over the total
        100 optimizer steps. Acc reports Same-Step Avg@4, averaging the last four
        evaluation checkpoints. Wall time sums per-step training
        time over all training steps.}
        \label{tab:abl}
        \scriptsize
        \def\interventionTableWidth{0.98\linewidth}
        \setlength{\tabcolsep}{3pt}
        \renewcommand{\arraystretch}{0.96}
        \resizebox{\interventionTableWidth}{!}{%
        \begin{tabular}{@{}>{\raggedright\arraybackslash}m{2.2cm}>{\centering\arraybackslash}m{1cm}>{\centering\arraybackslash}m{1cm}>{\centering\arraybackslash}m{1.2cm}>{\centering\arraybackslash}m{1.2cm}@{}}
        \toprule
        Config & Depth & Blend & Avg@4 & Wall (h) \\
        \midrule
        Vanilla OPD          &        &        & 83.0 & 4.42 \\
        + Adaptive Depth          & \checkmark &     & 82.8 & 1.96 \\
        + Linear blend norm         &        & \checkmark & 85.1 & 2.59 \\
        \textbf{TurnOPD}        & \checkmark & \checkmark & \textbf{86.3} & \textbf{1.93} \\
        \bottomrule
        \end{tabular}}
    \end{minipage}
\end{figure}

\textbf{Results.}
Figure~\ref{fig:ablation} presents the full validation curves across training, while Table~\ref{tab:abl} summarizes the Same-Step Avg@4 accuracy and cumulative wall time after 100 optimizer steps.
The results support the intended division of labor. Adaptive depth alone cuts
the 100-step wall time from $4.42$ h to $1.96$ h, but its accuracy drops slightly from
$83.0$ to $82.8$. This indicates that reducing rollout depth is not a complete
training solution by itself: it saves compute, but it does not fix the loss
allocation problem identified in Section~\ref{sec:diag:turnnorm}. In contrast,
the linear KL blend improves accuracy to $85.1$, showing that the loss
allocator directly improves optimization, but it still costs $2.59$ h. 
Full TurnOPD combines the two effects: it achieves the best average accuracy ($86.3$) while matching the low wall time of the adaptive-depth run ($1.93$ h). Notably, using adaptive rollout depth together with turn-level KL normalization alone can sometimes yield final performance close to, or slightly higher than, TurnOPD. However, TurnOPD outperforms these alternatives in the middle stages of training, demonstrating faster progress and better intermediate results. Thus, the ablations tell a coherent story: the rollout-depth budget controller provides the efficiency lever, the progressive loss-budget controller provides the optimization lever, and their combination gives the best accuracy--time tradeoff.

\subsection{Effect of KL Normalization}
\label{sec:exp:kl-normalization}

\textbf{Motivation and Experiment Setup.}
Section~\ref{sec:diag:turnnorm} points out that KL loss normalization has a budget allocation problem: the standard (trajectory-level) approach mostly focuses on shallow turns, while a hard turn-level strategy can overcompensate and put too much weight on rarely supported deep turns. To better understand this, we compare three KL normalization schemes: \textbf{trajectory-level KL}, \textbf{hard turn-level KL}, and our \textbf{linear blend method}. For fairness, we report each method's Same-Step Avg@4 and the actual loss budget spent on the deepest third of valid turns (i.e., turns reached by enough trajectories).

\textbf{Results.}
As shown in Table~\ref{tab:abl-turnnorm}, trajectory-level KL is stable but strongly favors shallow turns: the deepest third gets only 3.2/0.7/1.2\% of the budget in early/mid/late training. Hard turn-level KL sharply reverses this, instantly giving around a third of the budget to deep turns, but makes an abrupt shift and may overweight unreliable estimates early on. Our linear blend offers a smoother transition: $\alpha$ rises from $0.17$ to $0.83$ across the early/mid/late phases, while the deep-turn budget increases from $12.8\%$ to $27.7\%$. This method balances stability and targeted allocation, yielding the best Same-Step Avg@4, though the margin over hard turn-level KL is small.

\begin{table*}[htbp]
\centering
\caption{
Impact of KL normalization strategies on ALFWorld-1.7B. ``Same-Step Avg@4'' is the mean over the last four evaluation checkpoints. ``Deep budget'' gives the KL budget assigned to the deepest third of valid turns in early/mid/late training. ``$\alpha$'' is the blend coefficient, with $\alpha=0$ for trajectory-level KL and $\alpha=1$ for hard turn-level KL.
}
\label{tab:abl-turnnorm}
\small
\setlength{\tabcolsep}{4pt}
\resizebox{0.85\textwidth}{!}{%
\begin{tabular}{
    @{}>{\raggedright\arraybackslash}m{4.5cm}>{\centering\arraybackslash}m{2.2cm}>{\centering\arraybackslash}m{3.2cm}>{\centering\arraybackslash}m{2.2cm}@{}}
\toprule
KL normalizer & \shortstack{Same-Step\\Avg@4} & \shortstack{Deep budget\\E/M/L} & \shortstack{$\alpha$\\E/M/L} \\
\midrule
Trajectory-level KL          & 83.0 & 3.2/0.7/1.2\% & 0/0/0 \\
Turn-level KL (hard switch)  & 85.0 & 29.9/32.0/31.9\% & 1/1/1 \\
\textbf{Linear blend (ours)} & \textbf{85.1} & 12.8/21.6/27.7\% & 0.17/0.50/0.83 \\
\bottomrule
\end{tabular}
}
\end{table*}

\subsection{Adaptive-Depth Coverage Sensitivity}
\label{sec:exp:coverage-sensitivity}

The Adaptive Rollout Depth module exposes two main hyperparameters: (1) the coverage quantile $p$, and (2) the choice of whether the completion cumulative distribution function (CDF) is computed from only successful trajectories or from the full trajectory population. In practice, many open-ended agentic tasks do not provide a reliable distinction between successful and unsuccessful trajectories. Moreover, the OPD algorithm itself does not fundamentally depend on a reward signal denoting task success. To assess robustness in more general settings, we conducted additional experiments evaluating the adaptive-depth controller without access to a success-conditioned CDF.

For this hyperparameter study, we swept $p\in\{0.4, 0.6, 0.8\}$ across both success-conditioned and full-population CDFs. Figure~\ref{fig:coverage-ablation} summarizes four diagnostic metrics: validation accuracy and EMA-smoothed rollout horizon (ema-H) for both the success-conditioned and full-population CDF settings. Here, ema-H refers to the exponentially smoothed average rollout depth, which directly determines the effective student rollout horizon. We summarize the main observations as follows:

\begin{figure}[htbp]
\centering
\includegraphics[width=\linewidth]{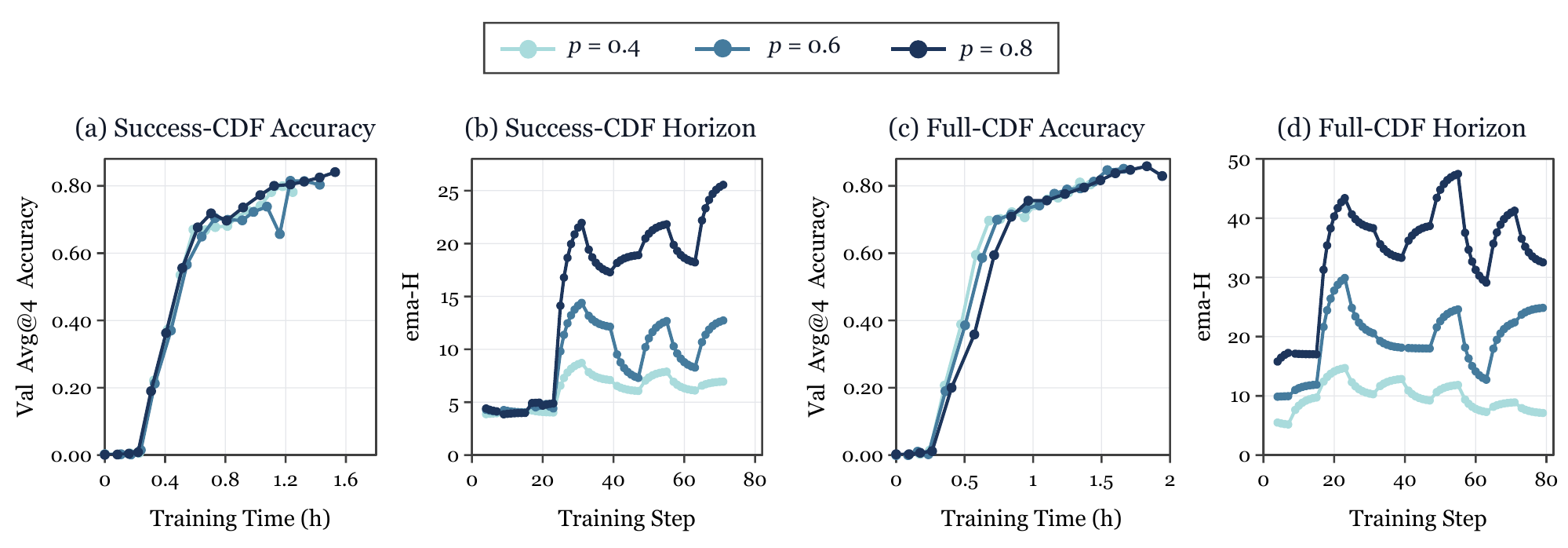}
\caption{Coverage-floor sensitivity on ALFWorld-1.7B. The four panels show validation accuracy versus
cumulative training time and logged EMA horizon for the success-conditioned CDF
and the full-population CDF.}
\label{fig:coverage-ablation}
\end{figure}

\textbf{(1) The quantile parameter $p$ directly controls the target rollout horizon.} Increasing $p$ yields slightly higher final success rates, but also results in deeper rollout horizons and correspondingly increased training time. For example, with the full-population CDF, setting $p=0.6$ achieves an accuracy of 85.1 in 1.66 hours, while $p=0.8$ improves the accuracy to 85.8 but increases the required time to 1.83 hours.

\textbf{(2) The CDF source significantly affects efficiency.} The full-population CDF is more aggressive: it consistently increases ema-H and wall time compared to the success-conditioned CDF. This matches intuition, as the full-population CDF incorporates many failed trajectories, including those that stall at the maximum rollout depth due to repeated or unproductive tool usage, thereby overestimating the required depth. Nonetheless, as shown in the figure, the full-population CDF remains effective, but requires appropriately lower values of $p$ to match the efficiency and performance of the success-conditioned CDF.

%% file: sections/conclusion.tex
\section{Conclusion}
\label{sec:limit}

We studied on-policy distillation for long-horizon language agent tasks. Our
diagnosis shows that vanilla OPD suffers from two allocation mismatches:
full-horizon rollouts can spend compute on low-yield tail turns, while
trajectory-level KL reduction can concentrate loss on shallow tokens. TurnOPD
addresses these mismatches with turn-level budgeting: it adapts rollout depth
and progressively shifts KL normalization toward turn-balanced supervision.
Across ALFWorld, WebShop, and Multi-Hop Search, TurnOPD advances the
accuracy--time frontier over vanilla OPD.
Overall, our results suggest a broader lesson for agent OPD. The unit of
supervision in long-horizon agents should not be treated as a flat token
position, but as a turn-conditioned decision inside an evolving interaction
trace. TurnOPD opens a path toward turn-aware and budget-adaptive training of
long-horizon language agents.


%% file: sections/appendix.tex
\newpage
\appendix


\section{Environments, Tool Configurations, And Task Examples}
\label{app:envs}

In this work, we use the following three environments and task types:
ALFWorld~\citep{shridhar2021alfworld}, Multi-Hop Search (consisting of several sub-benchmarks including PopQA~\citep{mallen2023popqa}, NQ~\citep{kwiatkowski2019natural}, 2WikiMultiHopQA~\citep{ho2020twowiki}, HotpotQA~\citep{yang2018hotpotqa}), and WebShop~\citep{yao2022webshop}.

For the student and teacher models, we explicitly activate the ``thinking'' mode, so each model turn consists of a segment of chain-of-thought and a tool call.
Client-side validation rejects malformed or inadmissible actions before they reach the
environment so a bad action does not consume a real step.
All experiments are conducted on 32 NVIDIA H20 GPUs, each with 80 GB of memory.

\subsection{Environment Details}

\subsubsection{ALFWorld (Embodied Household Planning)}
\label{app:envs:alfworld}

\textbf{Tool / action space.} A single tool \texttt{alfworld\_step(action: str)}; the
action is a natural-language household command from the templates: \texttt{go to
\{loc\}}, \texttt{take \{obj\} from \{loc\}}, \texttt{put \{obj\} in/on \{loc\}},
\texttt{open/close \{loc\}}, \texttt{use \{loc\}}, \texttt{clean/heat/cool \{obj\} with
\{loc\}}, \texttt{examine \{x\}}, \texttt{inventory}, \texttt{look}. The tool returns
the new TextWorld observation plus a refreshed list of admissible commands; actions not
in the admissible set are rejected client-side.

\textbf{Environment.} A FastAPI server wraps TextWorld~\citep{cote2018textworld} with a
spawned worker pool. The six task types are \texttt{pick\_and\_place},
\texttt{pick\_clean\_then\_place}, \texttt{pick\_heat\_then\_place},
\texttt{pick\_cool\_then\_place}, \texttt{look\_at\_obj\_in\_light}, and
\texttt{pick\_two\_obj\_and\_place}; evaluation uses the \texttt{valid\_unseen}
(out-of-distribution) split. The reward is \textbf{binary} ($1$ on goal completion,
else $0$).

\textbf{Task example.}
\begin{verbatim}
[system] You are an AI agent operating in a household environment. You can
interact with objects and navigate between locations using the alfworld_step tool.
[user]   Your task is to: put a clean mug in coffeemachine.
[assistant -> alfworld_step] {"action": "go to countertop 1"}
[tool]   On the countertop 1 you see a mug 1, a knife 1, ...
[assistant -> alfworld_step] {"action": "take mug 1 from countertop 1"}
[tool]   You pick up the mug 1 from the countertop 1.
[assistant -> alfworld_step] {"action": "clean mug 1 with sinkbasin 1"} ...
\end{verbatim}

\subsubsection{Multi-Hop Search}
\label{app:envs:searchr1}

\textbf{Tool / action space.} We instantiate Multi-Hop Search with a Search-R1-style interface \citep{jin2025searchr1}. A single retrieval tool
\texttt{search\_r1/search(query: str)} POSTs the query to a dense-retrieval server and returns the top-$k$ passages
formatted as \texttt{Doc i(Title: ...) ...}. The model
reasons in \texttt{<think>...</think>}, issues retrieval via the search tool (or the
equivalent \texttt{<search>...</search>} text channel), receives results in
\texttt{<information>...</information>}, and emits the final answer in
\texttt{<answer>...</answer>}.

\textbf{Environment.} The retrieval endpoint serves
a fixed Wikipedia corpus; rollouts are capped at \texttt{max\_turns}=50, allowing many
search--read hops. The reward is \textbf{exact match (EM)} of the extracted
\texttt{<answer>} against the gold answer.

\textbf{Sub-benchmarks.} The held-out Multi-Hop Search test set is a mixture of
four QA sources: \texttt{PopQA}~\citep{mallen2023popqa},
\texttt{Natural Questions} (NQ)~\citep{kwiatkowski2019natural},
\texttt{2WikiMultiHopQA}~\citep{ho2020twowiki}, and
\texttt{HotpotQA}~\citep{yang2018hotpotqa}. PopQA and NQ provide open-domain
retrieval questions, while 2WikiMultiHopQA and HotpotQA require chained
retrievals where a later query depends on an earlier passage, making the
evaluation mixture cover both direct retrieval and long-horizon dependency
cases.

\textbf{Task example.}
\begin{verbatim}
[user] Answer the question. Reason in <think>...</think>; if you lack knowledge,
search via <search> query </search> and read the returned <information>...
</information>; give the final answer in <answer>...</answer>.
Question: Who is the director of the film that won Best Picture in the year
Titanic was released?
[assistant] <think>I need the Best-Picture winner of 1997.</think>
            <search> 1997 Academy Award Best Picture winner </search>
[tool]      <information>Doc 1(Title: Titanic (1997 film)) ... won Best Picture ...
            </information>
[assistant] <think>Director of Titanic is James Cameron.</think>
            <answer> James Cameron </answer>
\end{verbatim}

\subsubsection{WebShop (Web Navigation / Online Shopping)}
\label{app:envs:webshop}

\textbf{Tool / action space.} The agent operates a single tool \texttt{webshop\_step(action: str)} with two formats: \texttt{search[<keywords>]} for query input (when a search bar is available), and \texttt{click[<button>]} to interact with one of the currently visible buttons (e.g., product ID, attribute options like ``x-large'', or navigation buttons such as \texttt{Buy Now}, \texttt{Next >}, etc.). The tool outputs the updated page view and available actions. Invalid clicks are blocked on the client side.

\textbf{Environment.} A FastAPI server wraps AgentGym's
\texttt{agentenv\_webshop}~\citep{xi2025agentgym}. An episode proceeds search $\to$ inspect results $\to$ open a
product $\to$ select attributes $\to$ \texttt{click[Buy Now]}. The reward is a
\textbf{dense terminal score in $[0,1]$} reflecting partial credit for matching the
requested product type, attributes, options, and price; only the terminating
\texttt{Buy Now} carries reward, and a machine-parseable
\texttt{[WEBSHOP\_FINAL\_SCORE=...]} marker is appended for the reward path.

\textbf{Task example.}
\begin{verbatim}
[user] Instruction: I am looking for a milk chocolate of 1 pound size in a single
pack for valentine day, and price lower than 40 dollars.
Available actions: search[<your keywords>]
[assistant -> webshop_step] {"action": "search[valentine milk chocolate 1 pound
single pack]"}
[tool]  [Results] B07... Valentine Milk Chocolate 1 lb ... $19.99  (clickables:
        "B07...", "Next >", "Back to Search")
[assistant -> webshop_step] {"action": "click[B07...]"}
[tool]  ... (clickables: "1 pound", "Buy Now", "Description", ...)
[assistant -> webshop_step] {"action": "click[Buy Now]"}
[tool]  Episode finished with score=1.000.  [WEBSHOP_FINAL_SCORE=1.000000]
\end{verbatim}

\subsection{Task-Specialized Teacher Training}
\label{app:teacher-grpo}

\begin{figure*}[htbp]
  \centering
  \includegraphics[width=\textwidth]{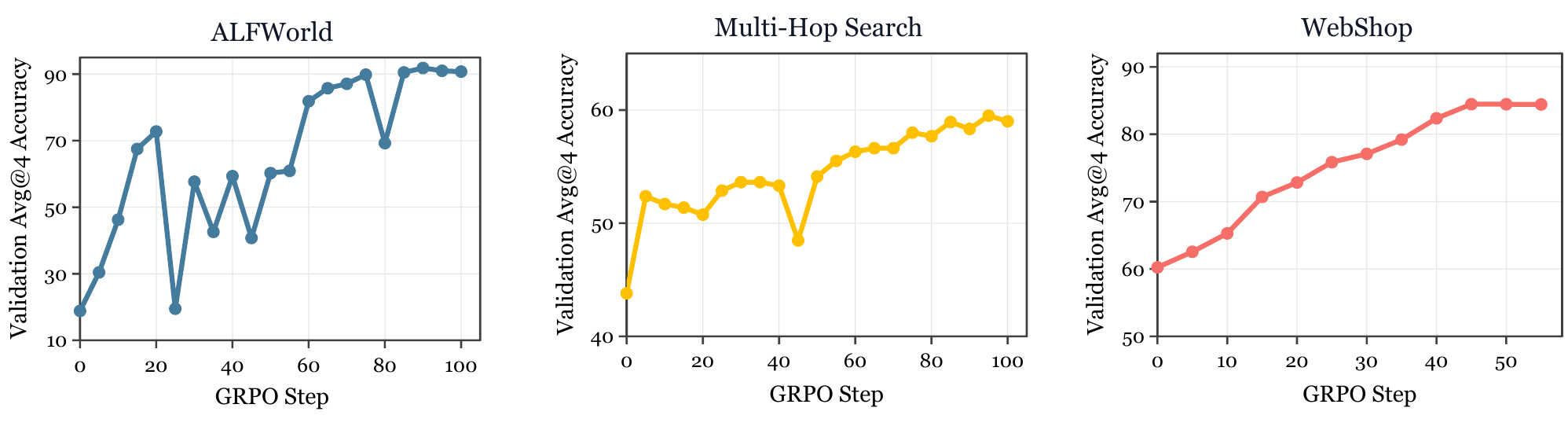}
  \caption{GRPO validation curves for the task-specialized teacher models.}
  \label{fig:teacher-grpo-val}
\end{figure*}

We train task-specialized teachers using GRPO~\citep{shao2024deepseekmath} on the three environments.
Figure~\ref{fig:teacher-grpo-val} reports the validation dynamics of the
task-specialized teachers used in the main experiments. All curves report avg@4 accuracy. ALFWorld and WebShop use Qwen3-8B-GRPO
teachers, while Multi-Hop Search uses a Qwen3.5-9B-GRPO teacher.

\section{Proof Of The Contamination-Compression Bound}
\label{app:contamination-proof}

We provide the full proof of Proposition~\ref{prop:contamination-compression}.
Fix a context $c$ and write $\lambda=\lambda(c)$ for readability. The only
ingredient is the joint convexity of KL divergence. For any distributions
$P_1,Q_1,P_2,Q_2$ over the same vocabulary and any $\alpha\in[0,1]$,
\begin{equation}
\KL\!\left(\alpha P_1+(1-\alpha)P_2\,\|\,\alpha Q_1+(1-\alpha)Q_2\right)
\le
\alpha\KL(P_1\,\|\,Q_1)+(1-\alpha)\KL(P_2\,\|\,Q_2).
\label{eq:app-joint-convex-kl}
\end{equation}

\textbf{Joint convexity from log-sum.}
For each token $x$, set
$a_1=\alpha P_1(x)$, $a_2=(1-\alpha)P_2(x)$,
$b_1=\alpha Q_1(x)$, and $b_2=(1-\alpha)Q_2(x)$. The log-sum inequality gives
\begin{align}
&\big(a_1+a_2\big)
\log\frac{a_1+a_2}{b_1+b_2} \nonumber\\
&\qquad\le
a_1\log\frac{a_1}{b_1}
  + a_2\log\frac{a_2}{b_2} \nonumber\\
&\qquad=
\alpha P_1(x)\log\frac{P_1(x)}{Q_1(x)}
  +(1-\alpha)P_2(x)\log\frac{P_2(x)}{Q_2(x)} ,
\end{align}
where the factors $\alpha$ and $1-\alpha$ cancel inside the logarithms. Summing
over all tokens $x$ gives Equation~\ref{eq:app-joint-convex-kl}.

\textbf{Applying the inequality.}
Set $\alpha=\lambda$, $P_1=Q_1=p_F$, $P_2=p_S^{\mathrm{free}}$, and
$Q_2=p_T^{\mathrm{free}}$. Then
\begin{align}
&\KL\!\left(
\lambda p_F + (1-\lambda)p_S^{\mathrm{free}}
\,\|\,
\lambda p_F + (1-\lambda)p_T^{\mathrm{free}}
\right) \nonumber\\
&\qquad\le
\lambda \KL(p_F\,\|\,p_F)
  + (1-\lambda)\KL(p_S^{\mathrm{free}}\,\|\,p_T^{\mathrm{free}}).
\end{align}
Because $\KL(p_F\,\|\,p_F)=\sum_x p_F(x)\log 1=0$,
\begin{equation}
\KL\!\left(
\lambda p_F + (1-\lambda)p_S^{\mathrm{free}}
\,\|\,
\lambda p_F + (1-\lambda)p_T^{\mathrm{free}}
\right)
\le
(1-\lambda)\KL(p_S^{\mathrm{free}}\,\|\,p_T^{\mathrm{free}}),
\end{equation}
which proves Equation~\ref{eq:contamination-bound}.

\section{Existence of a Latent Rollout-Depth Optimum}
\label{app:rollout-depth-existence}

This appendix justifies the notation $H^\star$ used in
Section~\ref{sec:method}. The useful rollout horizon is not defined by a single
force. It is squeezed between two opposing requirements. If the horizon is too
short, the rollout fails to cover the decision turns on which successful
trajectories finish. If the horizon is too long, the remaining turns add
collection cost after the measurable teacher-correction signal has already
decayed. This gives a two-sided efficiency--coverage sandwich.

\textbf{Efficiency side.}
Fix a student checkpoint, a task distribution, and a finite maximum rollout cap
$H_{\max}$. For an idealized depth $H$, let
\begin{equation}
\rho(H)
  =
  \frac{\sum_{t\le H} v_t w_t}
       {\sum_{t\le H} c_t w_t+\epsilon},
\label{eq:app-efficiency-rate}
\end{equation}
where $v_t\ge0$ is the local teacher-correction value at turn $t$,
$w_t=n_t/n_0$ is the survivor probability, and $c_t>0$ is the per-turn
collection cost. Let $H_{\mathrm{eff}}^\star$ be the smallest maximizer of
$\rho(H)$ on $1\le H\le H_{\max}$. Equivalently, adding turn $H+1$ improves the
ratio only when its marginal rate is at least the current average:
\begin{equation}
\rho(H+1)\ge \rho(H)
\quad\Longleftrightarrow\quad
\frac{v_{H+1}}{c_{H+1}}\ge \rho(H),
\label{eq:app-marginal-rate}
\end{equation}
obtained by cross-multiplying
$(A+a)/(B+b)\ge A/B$ for positive $B,b$. Suppose there is a signal-exhaustion
frontier $\tau_c$ after which the marginal teacher-correction value is zero, or
more generally the marginal rate stays below the running average. Then every
turn beyond $\tau_c$ can only decrease the efficiency ratio, so
\begin{equation}
H_{\mathrm{eff}}^\star \le \tau_c-1 .
\label{eq:app-heff-upper}
\end{equation}

\textbf{Coverage side.}
Let $L_{\mathrm{succ}}$ be the completion length of a successful rollout, and
let
\begin{equation}
F_{\mathrm{succ}}(H)=\Pr(L_{\mathrm{succ}}\le H),\qquad
H_{\mathrm{cov}}=\widehat Q_p(L_{\mathrm{succ}})
  =\min\{H:F_{\mathrm{succ}}(H)\ge p\},
\label{eq:app-hcov}
\end{equation}
for a fixed $p\in(0,1]$. This is a coverage floor: stopping before
$H_{\mathrm{cov}}$ misses the completion decisions of more than a $(1-p)$
fraction of successful trajectories. Define the full-coverage completion depth
as
\begin{equation}
\tau_{\mathrm{done}}
  =
  \min\{H:F_{\mathrm{succ}}(H)=1\}.
\label{eq:app-tau-done}
\end{equation}
Equivalently, for a finite empirical support this is the largest observed
successful completion length, $\max\mathrm{supp}(L_{\mathrm{succ}})$. This
definition is deliberately the first point that has \emph{already} covered all
successful completions, not the last point before full coverage. Therefore
\begin{equation}
H_{\mathrm{cov}}\le \tau_{\mathrm{done}} .
\label{eq:app-hcov-upper}
\end{equation}

\textbf{Sandwich.}
The latent target should respect both sides:
\begin{equation}
H^\star
  =
  \max\!\left(H_{\mathrm{eff}}^\star,\ H_{\mathrm{cov}}\right).
\label{eq:app-hstar-sandwich-def}
\end{equation}
Therefore
\begin{equation}
H_{\mathrm{eff}}^\star
\le
H^\star
\le
\max\!\left(\tau_c-1,\tau_{\mathrm{done}}\right).
\label{eq:app-hstar-sandwich}
\end{equation}
The lower side prevents an overly shallow efficiency optimum from truncating
turns that successful trajectories need; the upper side prevents the horizon
from extending past both the signal-exhaustion frontier and the deepest
successful completion.

\section{Complete TurnOPD Hyperparameters}
\label{app:hyper}

Table~\ref{tab:hyper-full} lists the full configuration of the main TurnOPD method,
mapping each implementation key to the symbol used in the formal model
(Sections~\ref{sec:method:A}--\ref{sec:method:B}).
The \texttt{ema\_alpha} parameter is the EMA weight on the depth proxy
$H_{\mathrm{ctrl}}$, distinct from the loss-blend coefficient $\alpha$.

\begin{algorithm}[H]
  \small
  \caption{TurnOPD training algorithm.}
  \label{alg:tasc}
  \DontPrintSemicolon
  \KwIn{$\pi_\theta$, $\pi_T$, task distribution $\mathcal D$, environment turn limit
  $T_{\max}$, cap bounds $H_{\min},H_{\max}$, training horizon $K$, and controller hyperparameters
  $K_{\mathrm{warm}},r_{\mathrm{probe}},\alpha_{\mathrm{ema}},
  p,n_{\min}^{\mathrm{cov}},(s,e)$.}
  \KwOut{updated student policy $\pi_\theta$.}
  \BlankLine
  Initialize $\bar H \leftarrow \varnothing$ (unset) and $\Hcov \leftarrow 0$\;
  \For{$k=1$ \KwTo $K$}{
    $\mathrm{probe}_k \leftarrow (k\le K_{\mathrm{warm}})\ \lor\
    (k\bmod r_{\mathrm{probe}}=0)$\;
    \eIf{$\mathrm{probe}_k$}{
      $H_{\mathrm{roll}} \leftarrow T_{\max}$\tcp*[r]{full-depth probe, no adaptive cap}
    }{
      $\hat H_k \leftarrow
      \mathrm{clip}(\mathrm{round}(\bar H)+1,H_{\min},H_{\max})$\;
      $H_{\mathrm{roll}} \leftarrow \hat H_k$\;
    }
    Collect on-policy trajectories $\mathcal B_k$ from $\pi_\theta$ on
    $\mathcal D$ up to $H_{\mathrm{roll}}$ turns\;
    Query $\pi_T$ on the supervised tokens in $\mathcal B_k$ and compute
    top-$K$ reverse-KL token losses $\ell_i$\;
    Compute per-turn raw-KL means $K_t$ and survivor counts $n_t$ on
    $\mathcal B_k$\;
    $m_t \leftarrow [K_t]_+\,n_t/n_0$,\quad
    $q_t \leftarrow m_t/(\sum_j m_j+\epsilon)$\;
    $\Heff \leftarrow \mathrm{round}(\sum_t t\,q_t)$\;
    \If{$\mathrm{probe}_k \land |\mathcal B_k^{\mathrm{succ}}|\ge n_{\min}^{\mathrm{cov}}$}{
      Estimate $F_{\mathrm{succ}}(H)$ from successful probe trajectories and set
      $\Hcov \leftarrow \min\{H:F_{\mathrm{succ}}(H)\ge p\}$\;
    }
    $H_{\mathrm{ctrl}} \leftarrow \max(\Heff,\Hcov)$\;
    \eIf{$\bar H=\varnothing$}{
      $\bar H \leftarrow H_{\mathrm{ctrl}}$\tcp*[r]{initialize from the first target}
    }{
      $\bar H \leftarrow (1-\alpha_{\mathrm{ema}})\bar H
      +\alpha_{\mathrm{ema}}H_{\mathrm{ctrl}}$\;
    }
    $\rho_k \leftarrow k/K$,\quad
    $\alpha_k \leftarrow \mathrm{clip}((\rho_k-s)/(e-s),0,1)$\;
    Construct trajectory-normalized weights $w_{\mathrm{traj}}$ and
    turn-normalized weights $w_{\mathrm{turn}}$ on the observed tokens in
    $\mathcal B_k$\;
    $w_{\mathrm{blend}} \leftarrow
    (1-\alpha_k)w_{\mathrm{traj}}+\alpha_k w_{\mathrm{turn}}$\;
    $\mathcal L_k \leftarrow \sum_{i\in\mathcal B_k}
    w_{\mathrm{blend},i}\ell_i$\;
    Update $\theta$ by descending $\nabla_\theta \mathcal L_k$\;
  }
  \end{algorithm}

\begin{table*}[tbp]
\centering
\caption{Full TurnOPD configuration (ALFWorld-1.7B reference run). ``Symbol'' is the
variable in the formal model; ``Key'' is the implementation config name.}
\label{tab:hyper-full}
\scriptsize
\setlength{\tabcolsep}{3.5pt}
\resizebox{\textwidth}{!}{%
\begin{tabular}{@{}>{\centering\arraybackslash}m{2.8cm}>{\centering\arraybackslash}m{2.2cm}>{\centering\arraybackslash}m{5.3cm}>{\centering\arraybackslash}m{6.6cm}@{}}
\toprule
Group & Symbol & Key = value & Role \\
\midrule
\multirow{3}{*}{External signal}
 & $K_t$ & \texttt{kl\_per\_turn/*} & raw per-turn reverse-KL signal used in $m_t=[K_t]_+n_t/n_0$ \\
 & $n_t$ & \texttt{num\_traj/*} & all-trajectory survivor count used in the efficiency-side mass \\
 & $K_{\mathrm{top}}$       & \texttt{distill\_topk}=50 & teacher top-$K_{\mathrm{top}}$ support size for the reverse-KL loss \\
\midrule
\multirow{4}{*}{External depth}
 & $p$ & \texttt{coverage\_quantile}=0.80 & success-length quantile for $\Hcov=\hat Q_p(L_{\mathrm{succ}})$ \\
 & --- & \texttt{use\_success}=True & read $\Hcov$ from success subset (False $\to$ all-trajectory CDF) \\
 & --- & \texttt{min\_cov\_traj}=8 & min successful trajectories to refresh $\Hcov$ \\
 & $H_{\min}/H_{\max}$ & \texttt{min/max}=2/50 & clamp on applied depth $\hat H$ \\
\midrule
\multirow{3}{*}{External online}
 & $\alpha_{\mathrm{ema}}$ & \texttt{ema\_alpha}=0.30 & EMA weight on $H_{\mathrm{ctrl}}$ (not the blend $\alpha$) \\
 & --- & \texttt{probe\_interval}=8 & steps between full-depth probes \\
 & --- & \texttt{warmup\_steps}=3 & probe-only warmup before truncation \\
\midrule
\multirow{4}{*}{Internal blend}
 & $s,e$ & \texttt{blend start/end}=0/1 & linear progress window for $\alpha=\mathrm{clip}((\mathrm{prog}-s)/(e-s),0,1)$ \\
 & --- & \texttt{min\_floor}=8 & turn-level $n_{\min}=\max(\text{floor},\lceil\text{frac}\cdot n_{\mathrm{traj}}\rceil)$ \\
 & --- & \texttt{min\_frac}=0.15 & fraction term of $n_{\min}$ \\
 & --- & \texttt{turn\_norm\_blend}=True & enable linear trajectory$\to$turn blend \\
\midrule
\multirow{6}{*}{OPD / optimization}
 & --- & \texttt{kl\_coef}=0, \texttt{kl\_loss\_coef}=0 & no reward/penalty KL; loss is pure top-$K$ RKL \\
 & --- & \texttt{loss\_mode}=topk\_reverse\_kl & differentiable objective \\
 & --- & \texttt{seq-mean-token-mean} & base (trajectory-level) aggregation \\
 & --- & \texttt{clip\_ratio}=0.9/9 & importance-ratio clip (low/high) \\
 & --- & lr=$10^{-6}$, batch=64, steps=100 & AdamW \\
\midrule
\multirow{3}{*}{Rollout / eval}
 & --- & \texttt{max\_turns}=80 (ALFWorld) / 50 (Multi-Hop Search) & rollout turn cap before adaptive-depth truncation \\
 & --- & temperature=1.0, $n$=1 & training rollout sampling \\
 & --- & \texttt{val\_n}=4, val temp.=0.85 & validation: mean@4 \\
\bottomrule
\end{tabular}
}
\end{table*}

\FloatBarrier

\section{OPD Data Scale And Train/Test Construction}
\label{app:data}

Each task feeds the same on-policy distillation pipeline a parquet of prompts; the
agent then rolls out against the live environment, so a ``sample'' is a task
\textbf{instance} (a goal / question) rather than a fixed teacher trajectory---trajectories
are generated online during training. Table~\ref{tab:data} summarizes the reference
scale and split construction; the three tasks differ fundamentally in how their
train/test partitions are obtained, which we detail below.

\begin{table*}[tbp]
\centering
\caption{OPD dataset scale and split construction. ``Train/Eval'' counts are task
instances (online rollouts, not stored trajectories); the per-trajectory step cap is the
data-side \texttt{max\_iterations} before any adaptive-depth truncation.}
\label{tab:data}
\small
\setlength{\tabcolsep}{8pt}
\resizebox{\textwidth}{!}{%
\begin{tabular}{@{}>{\centering\arraybackslash}m{3.0cm}>{\centering\arraybackslash}m{2.2cm}>{\centering\arraybackslash}m{2.0cm}>{\centering\arraybackslash}m{6.8cm}@{}}
\toprule
Task & Train & Eval & Eval--train relation \\
\midrule
ALFWorld   & $\approx5000$ & $\approx300$ & held-out OOD split (\texttt{eval\_out\_of\_distribution}) \\
Multi-Hop Search  & $\approx30000$ & $\approx400$ & separate four-source QA test mixture \\
WebShop    & $\approx3000$ & $\approx200$ & disjoint goal-id ranges \\
\bottomrule
\end{tabular}
}
\end{table*}

\textbf{ALFWorld (procedurally generated, OOD evaluation).} Task instances are
synthesized by sampling one of the six task types uniformly
(Appendix~\ref{app:envs:alfworld}) and instantiating it with an object--receptacle pair
drawn from fixed vocabularies under a seeded RNG, so train and validation prompts are
reproducible and non-overlapping by seed. Crucially the two splits are routed to
\textbf{different} environment partitions: training instances use the \texttt{train}
games, whereas evaluation instances use the \texttt{eval\_out\_of\_distribution}
(valid-unseen) games, so reported accuracy measures generalization to unseen household
layouts rather than memorization. The reference configuration generates $\approx5000$
training and $\approx300$ evaluation instances with a per-trajectory step cap of $80$;
the underlying ALFWorld task pool has roughly $3.5\mathrm{k}$ unique games, which upper-bounds
the distinct training layouts.

\textbf{Multi-Hop Search (external QA mixture, separate test set).} Training uses an
externally prepared corpus of short-chain multi-hop / open-domain questions merged from
Search-R1-format sources, at $\approx30\mathrm{k}$ question--answer instances;
evaluation uses a separate held-out test set ($\approx400$ questions) drawn from
PopQA, Natural Questions (NQ), 2WikiMultiHopQA, and HotpotQA
(Appendix~\ref{app:envs:searchr1}), with the originating source preserved in the
\texttt{data\_source} field so that exact-match reward and per-source breakdowns route
correctly. Prompts are prepared with thinking enabled (a \texttt{<think>} reasoning
channel before each search), and the step cap is $50$ to accommodate several
search--read hops. Because the questions are real (not synthesized), train and test are
disjoint by construction at the dataset level.

\textbf{WebShop (goal-id partition, disjoint ranges).} A WebShop instance is selected
purely by a goal id (forwarded to the environment as the session/goal index); the
instruction text itself is part of the environment's initial observation and is not
duplicated into the prompt. Train and evaluation use \textbf{strictly disjoint} goal-id
ranges---evaluation takes ids $[0, N_{\mathrm{eval}})$ and training takes
$[N_{\mathrm{eval}}, N_{\mathrm{eval}}+N_{\mathrm{train}})$---so no goal appears in both
splits. The reference configuration uses $\approx3000$ training and $\approx200$
evaluation goals with a step cap of $30$ (WebShop episodes are short, typically 4--12
actions). Goal ids must stay within the goal count of the launched product corpus.